\documentclass[conference]{IEEEtran}
\IEEEoverridecommandlockouts

\IEEEpubid{\makebox[\columnwidth]{978-1-7281-8393-0/21/\$31.00~\copyright2021 IEEE \hfill} \hspace{\columnsep}\makebox[\columnwidth]{ }}

\usepackage{cite}
\usepackage{amsmath,amssymb,amsfonts}
\usepackage{algorithm}
\usepackage[noend]{algpseudocode}

\usepackage{graphicx}
\graphicspath{{figures/}}
\usepackage{textcomp}
\usepackage{xcolor}
\usepackage{booktabs}
\usepackage{subfigure}

\def\BibTeX{{\rm B\kern-.05em{\sc i\kern-.025em b}\kern-.08em
    T\kern-.1667em\lower.7ex\hbox{E}\kern-.125emX}}
\usepackage{empheq}
\usepackage{flushend}

\usepackage{bm}
\usepackage{stfloats}
\let\vec\mathbf
\begin{document}

\title{Attention-oriented Brain Storm Optimization for Multimodal Optimization Problems\\
  \thanks{This work is partially supported by the Science and Technology
    Innovation Committee Foundation of Shenzhen under the Grant
    No. JCYJ20200109141235597 and ZDSYS201703031748284, National Science
    Foundation of China under grant number 61761136008, Shenzhen
    Peacock Plan under Grant No. KQTD2016112514355531, and Program for
    Guangdong Introducing Innovative and Entrepreneurial Teams under
    grant number 2017ZT07X386.}}

\author{\IEEEauthorblockN{Jian Yang}
\IEEEauthorblockA{\textit{Department of Computer Science and Engineering} \\
\textit{Southern University of Science and Technology}\\
Shenzhen, China \\
yangj33@sustech.edu.cn}
\and
\IEEEauthorblockN{Yuhui Shi}
\IEEEauthorblockA{\textit{Department of Computer Science and Engineering} \\
\textit{Southern University of Science and Technology}\\
Shenzhen, China \\
shiyh@sustech.edu.cn}

}

\maketitle


\IEEEpubidadjcol

\begin{abstract}
  Population-based methods are often used to solve multimodal optimization problems. By combining niching or clustering strategy, the state-of-the-art approaches generally divide the population into several subpopulations to find multiple solutions for a problem at hand. However, these methods only guided by the fitness value during iterations, which are suffering from determining the number of subpopulations, i.e., the number of niche areas or clusters. To compensate for this drawback, this paper presents an Attention-oriented Brain Storm Optimization (ABSO) method that introduces the attention mechanism into a relatively new swarm intelligence algorithm, i.e., Brain Storm Optimization (BSO). By converting the objective space from the fitness space into ``attention'' space, the individuals are clustered and updated iteratively according to their salient values. Rather than converge to a single global optimum, the proposed method can guide the search procedure to converge to multiple ``salient'' solutions. The preliminary results show that the proposed method can locate multiple global and local optimal solutions of several multimodal benchmark functions. The proposed method needs less prior knowledge of the problem and can automatically converge to multiple optimums guided by the attention mechanism, which has excellent potential for further development.
\end{abstract}


\section{Introduction}

Population-based algorithms, including Evolutionary Algorithms (EAs) and Swarm Intelligence Algorithms (SIAs), are originally designed to locate a single global optimal solution for specific optimization problems. Under the selection pressure brought by fitness evaluation, these algorithms greedily converge to a single optimal. Nevertheless, many practical optimization problems contain multiple optimal solutions, such as electromagnetic design \cite{yoo2017new}, protein structure prediction \cite{nazmul2020multimodal}, power system design \cite{goharrizi2015parallel}, multi-target search in swarm robotics \cite{yang2020exploration}, as well as water supply networks \cite{yan2019multimodal}. These problems are known as multimodal optimization problems (MMOPs), which are required to find multiple global or local optimal in the solution space. It is a challenging task in the domain of optimization.

Many improvements have been proposed to allow the original population-based optimization algorithm to have multimodal optimization capabilities. These methods use the niching or clustering methods to divide the population into sub-populations to locate multiple global or local optimal solutions \cite{dai2021optimaidentified}. Niching methods maintain the formation of multiple sub-populations in niches and can be incorporated into standard population-based algorithms to solve the MMOPs. The niching strategies in literature include fitness sharing \cite{goldberg1987genetic, della2007niches}, crowding \cite{mahfoud1992crowding}, clearing \cite{petrowski1996clearing}, derating \cite{beasley1993sequential}, tournament selection \cite{harik1995finding}, speciation \cite{hui2015ensemble}, and parallelization \cite{bessaou2000island}, etc. The clustering approaches group individuals into clusters according to their similarity. Clustering-based methods divide the population into different clusters in every generation, increasing the likelihood of maintaining multiple optimal solutions. The Brain Storm Optimization originally has the clustering operation, which can be naturally applied to multimodal optimization problems. Other EAs or SIAs incorporated with clustering methods can be found in literature such as employing standard k-means into Particle Swarm Optimization (kPSO) \cite{passaro2008particle}, introducing affinity propagation clustering method into differential evolution \cite{wang2019automatic}, combining the clustering-based crowding and speciation with ant colony optimization \cite{yang2016adaptive}, etc.

However, these methods only guided by the fitness value during iterations, which are suffering from determining the number of subpopulations, i.e., how many niches or clusters are needed for a particular multimodal problem. One can not know how many optimal solutions exist for many problems, so we cannot divide the subpopulation in advance. Therefore, in niching or clustering technics, the corresponding adaptive mechanism is needed \cite{passaro2008particle}. Moreover, for clustering-based methods, how to define the similarity metrics is particularly important. The spatial information in solution space can indicate the distance between individuals but lack the fitness similarity. Clustering in objective space, i.e., using the fitness information for population partition, can guide the algorithm to locate the global optimal but cannot find more local optimal.

What is illustrated in this paper is an Attention-oriented BSO (ABSO) algorithm for multimodal optimization problems. The proposed method adopts the attention mechanism and converts the fitness space to the ``attention'' space for clustering and new solution generating, further guiding the algorithm to multiple ``salient'' solutions. The human attention mechanism can quickly select relatively important information for priority processing. For example, in the human visual perception system, there is far more information reaching onto our retinas than our cognitive systems at one time. Humans and other animals have evolved a remarkable attention mechanism that can select several salient cues from extensive visual information for brain processing, resulting in a good performance on dealing with complex dynamic scenes \cite{yang2020visual}. This mechanism has been used in many applications such as image processing \cite{yang2015multiresolution}, surveillance system \cite{jian2015cloud,yang2018line}, and other machine learning applications \cite{vaswani2017attention}, etc.

In the attention mechanism, the more ``salient'' part in massive information will attract more attention. The salient degree depends on the differences between it and its neighboring information, which refers to the center-surrounding principle \cite{borji2012state}. In multimodal optimization problems, the local optimal are more ``salient'' from its surroundings. Introducing the attention mechanism can guide the searching process to converge to multiple salient parts, which meet the requirements of MMOPs. By combining this mechanism with the BSO algorithm, we emphasize the following contributions of this paper: 1) introducing the attention mechanism into population-based algorithms for MMOPs. 2) converting the fitness value space into the ``attention'' space, in which the clustering operation is performed. 3) An archiving strategy is designed not only to store potential solutions but also to retain the global optimization capability.

The remainder of this paper is arranged as follows. Section 2 reviews the related works, including the BSO-related algorithms and other strategies, such as archiving and redistribution operations, for multimodal optimization problems. Section 3 gives the proposed ABSO algorithm, including the general procedure and other strategies used in this method. Section 4 presents the results on several representative benchmark functions with the convergence process. The discussion and conclusion are presented in Section 5 and Section 6, respectively.

\section{Related works}
This section will introduce the related works, including the Brain Storm Optimization algorithm and its variants and applications in multimodal optimization problems. Furthermore, other commonly used strategies in multimodal optimization problems, such as archiving and redistribution, are also briefly summarized.
\subsection{Brain Storm Optimization}
Brain storm optimization (BSO) algorithm is a relatively new swarm intelligence algorithm based on a collective human activity, i.e., the brainstorming process. Through convergence and divergence operations iteratively, a ``good enough'' optimal solution can be found in the search space. There are three types of operations in the original BSO algorithm: clustering, new individuals generation, and selection \cite{shi2011Optimization}.

The purpose of clustering is to converge the individuals into several smaller regions. Different clustering algorithms can be used \cite{cheng2017comprehensive}. The original BSO algorithm adopted the k-means clustering. Many BSO varieties have modified the clustering strategies to speed up the clustering process or change the number of cluster adaptively during iteration, such as the simple grouping method \cite{zhan2012modified}, random grouping strategy \cite{cao2015random}, max-fitness clustering \cite{guo2014modified}, affinity propagation clustering \cite{chen2015enhanced}, etc.

During iteration, a new individual can be generated based on one or several cluster(s). Generating an individual from one cluster can refine the search area and improve exploitation capabilities. Conversely, individuals generated from two or more clusters may be far away from these clusters, so the exploration ability is enhanced. The new individual can be generated by adding Gaussian random values to a selected old idea according to (\ref{eq:disOne}):
\begin{equation}
  \label{eq:disOne}
  x_{new}^i = x_{old}^i+\xi(t)\times rand()
\end{equation}
where $x_{new}^i$ represents the $i$th dimension of the newly generated idea, and $x_{old}^i$ is the selected old individual, which can be represented in (\ref{eq:base}), $rand()$ is a random function to generate a uniformly distributed random number in $[0,1)$. The coefficient $\xi(t)$ in (\ref{eq:disOne}) is the step size which weights the contribution of the Gaussian random value to the new generated value, which can be obtained according to (\ref{eq:xi}):
\begin{equation}
  \label{eq:xi}
  \xi(t) = logsig(\frac{\frac{T}{2}-t}{k})rand()
\end{equation}
where $logsig()$ is a logarithmic sigmoid transfer function, $T$ is the predefined maximum number of iterations, $t$ is the current iteration number, and $k$ is for changing $logsig()$ function's slope.
\begin{subequations}
  \label{eq:base}
    \begin{empheq}[left={x_{old}^i=\empheqlbrace\,}]{align}
      & x_{old1}^i \label{eq:base:a} \\
      & r(t) x_{old1}^i + (1-rand())x_{old2}^i \label{eq:base:b}
    \end{empheq}
\end{subequations}
where $x_{old1}$ and $x_{old2}$ represent individuals selected from two clusters. (\ref{eq:base:a}) is used to obtain the old idea based on one idea selected from one cluster, and (\ref{eq:base:b}) is used to obtain the old idea based on two ideas selected from the two clusters. For the new solution generation operation, there are also a series of improvements to the original BSO, such as modified search step size \cite{zhou2012brain}, discussion mechanism \cite{yang2015advanced}, learning strategy \cite{shen2020bso,qu2020bso}, etc.

The selection strategy in BSO is to keep good solutions in all individuals. The better solution is kept by the selection strategy after each new individual generation, while clustering strategy and generation strategy add new solutions into the swarm to keep the diversity for the whole population.

It's very worth mentioning that the BSO in Objective Space (BSO-OS) used the fitness values for clustering, i.e., the population is grouped into elite and normal clusters in each iteration \cite{shi2015brain}. Because the computation is only in objective space, it significantly reduced computing costs, and also has the convergence and divergence operations. Based on this clustering method, the corresponding new solution generation operation is slightly different from the original BSO algorithm. The main difference lies in the choice of $x_{old1}^i$ and $x_{old2}^i$ in (3). Because this clustering method divides all individuals into two categories, it selects seed solution from either elite class or normal class to generate new solutions. Other operations are consistent with the original BSO. The procedure of the BSO-OS is as shown in Alg.\ref{alg:bso}.

\begin{algorithm}[!htb]
  \caption{The BSO-OS Procedure} \label{alg:bso}
  \begin{algorithmic}[1]
    \State Randomly generate $n$ potential solutions (individuals);

    \State Evaluate the generated $n$ solutions;

    \While{not terminated}

    \State \textbf{Clustering}: Taking top $perc_e$ percentage as elitists and remaining
as normals;
    
    \State \textbf{New individuals generation}: Randomly select one or two individuals from either elitists or normals to generate $n$ new individuals;
    
    \State \textbf{Selection}: The fitness values of the newly generated individuals are compared with the existing individuals with the same index, the better one is kept;
    \EndWhile
  \end{algorithmic}
\end{algorithm}

\subsection{The BSO in Multimodal Optimization}
Multimodal optimization problems require finding as many global or local optimal solutions as possible in the solution space and saving the found optima in the iterative process. There are many attempts to apply the BSO algorithm to multimodal optimization problems. For example, the BSO-OS was applied to a dynamic multimodal problem to search and maintain multiple optimal during iterations \cite{cheng2018bicta}. The self-adaptive BSO (SBSO), which adopts a max-fitness clustering method (MCM), was proposed to solve MMOPs \cite{guo2014modified}. Further, a method of adaptive control the number of clusters for MCM was then presented \cite{dai2019modified}. Using MCM as well, an optima-identified framework combined with BSO (OIF-BSO) was proposed for this kind of problem \cite{dai2021optimaidentified}. This method combines disruption and redistribution strategies to locate and maintain the global optimum. The crowding strategy and roulette wheel selection methods are also applied to enhance local search capabilities.
\subsection{Other Strategies for MMOP}
In addition to the previously mentioned niching and clustering strategies, other strategies are often used in solving MMOPs. One of the representative strategies is archiving. This strategy is mainly used to maintain the optimal that have been found during the iteration. There are many archiving strategies currently used in MMOPs. For example, store potential optimal into a dynamic archive during iteration with \cite{lung2007new} or without \cite{epitropakis2013dynamic} refining operations. Another typical operation in MMOP is redistribution, that is, redistribute part of the population when appropriate. It can increase the diversity of the population and can avoid unnecessary evaluations around some identified peaks.
\section{Attention-oriented BSO}
As mentioned earlier, we will use the attention mechanism to operations of clustering and archiving. Besides, a redistribution strategies will be introduced to increase the diversity of the population in the iterative process. Details are as follows:
\subsection{Procedure of ABSO}
\begin{figure}[!htb]
  \centering
  \includegraphics[width=0.5\textwidth]{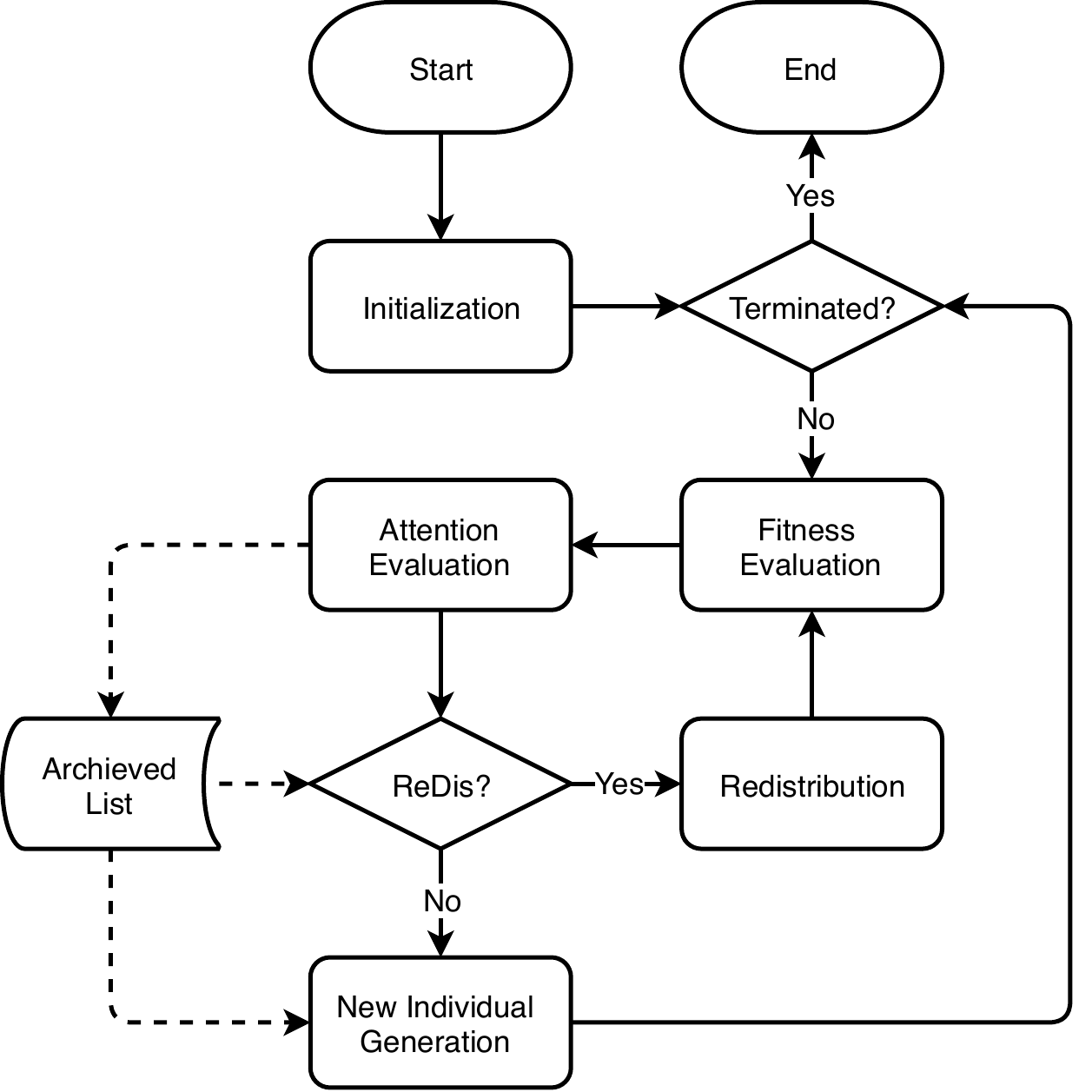}
  \caption{The Flowchart of ABSO}
  \label{fig:flowchart}
\end{figure}

The overall procedure of the proposed method is shown in Fig.\ref{fig:flowchart}. After initializing the population, the following steps are executed iteratively if the termination condition is not reached. First, evaluate each individual's fitness value in the population and then evaluate each individual's saliency value. Next, the solution that meets the pre-defined Archive condition is stored in the Archived list. If the redistribution condition is not met, a new solution is generated based on the attention-oriented clustering. On the contrary, if the redistribution condition is met, the new solution after redistribution will be evaluated with the fitness and attention evaluations, then continue the subsequent process. The archived list will be involved in operations of determining whether to perform the redistribution and new solutions generation. The final output of the algorithm is also this archived list.

\subsection{Attention Mechanism for BSO}
The attention mechanism describes that an event that is quite different from the surrounding will attract more attention, i.e., the center-surrounding principle \cite{borji2012state}. Generally, the degree of the bottom-up attention (non-subjective) can be evaluated by the saliency value of an event. According to the above principle, this paper uses (\ref{eq:saliency}) to define the saliency value $s$ of an individual $\vec{x}$:
\begin{equation}
  \label{eq:saliency}
  s_i = \frac{1}{m}\sum_{j=1}^mw_{ij}[f(\vec{x}_i)-f(\vec{x}_j)]
\end{equation}
where $f(\vec{x}_i)$ is the fitness value of $\vec{x}_i$, $m$ is the number of nearest individuals around $\vec{x}_i$, which can be the number of individuals within a certain radius around $\vec{x}_i$, or can be predefined to use $m$ nearest individuals around $\vec{x}_i$. This paper uses the KNN algorithm to find the nearest $m$ individuals in the current generation population and archived list to calculate the saliency. $w_{ij}$ is the weight ensures that farther individuals contribute less to the saliency value of the current one. It can be determined by (\ref{eq:weight}):
\begin{equation}
  \label{eq:weight}
  w_{ij} = \text{exp}\frac{-d_{ij}}{\sigma^2}
\end{equation}
where $d_{ij}$ is the euclidean norm between individual $i$ and $j$, $\sigma$ is the attenuation coefficient, which is the memic of the center-surrounding principle in the attention mechanism. This saliency value represents an individual's significance relative to surroundings. Note that the definition here is different for different problems. If the problem is to find the maximum value(s), the larger the value, the more salience it is. Conversely, if it is a minimization problem, the smaller the value, the more salience it is. Furthermore, for maximization problems, if $s_i>0$, we call $\vec{x}_i$ is a relatively salient solution ($s_i<0$ for minimization problems). The corresponding concept is absolutely salient solution, which will be introduced below. In addition, since we will use the attention mechanism to guide the search process, i.e., the objective function will also change to (\ref{eq:saliency}).

In the BSO-OS, the individuals are classified into elites and normals in the fitness value space. Here we will continue to use this clustering method, but it will be done in the ``attention'' space, i.e., the individuals with top $perc_e$\% in salient value will be clustered into a salient class, and others are clustered into a non-salient class. The salient individuals may be far away in the solution space, i.e., there are several salient individuals in the solution space, which aligns with the human brain's parallel processing of multiple cues (the multi-focal attention) \cite{mcmains2004multiple}. Under the guidance of this method of clustering, new solutions can be generated. Naturally, by searching around relatively salient individuals, a more salient solution can be found with a greater probability. Meanwhile, to improve the diversity of the algorithm, searching around non-salient solutions can extend exploration capabilities.

Besides, since we do not directly use fitness to guide the search process but apply the clustering and new solution generation in the ``attention'' space, the global optimization ability of the algorithm may be affected. Therefore, when calculating the saliency value of an individual through (\ref{eq:saliency}), both the individuals of the current generation and the solution that has been stored in the archive will participate in the calculation. According to the archiving strategy defined below, archiving solutions will be become better and better during iteration, thus ensuring the global optimization ability of the algorithm.

\subsection{Archiving Strategy}
The archiving strategy aims to store the possible optimal solutions to prevent the lost of good solutions in subsequent iterations. What kind of solutions will be stored to the list is the key to designing this strategy. In the proposed method, we define the concept of absolute saliency, i.e., for maximization problems, when an individual's fitness value is greater (or smaller for minimization problems) than the fitness value of all its surrounding solutions in the population, the individual is called an absolutely salient solution, as shown in (\ref{eq:abs}). When a solution meets the above definition, it will be put into the archived list.
\begin{equation}
  \label{eq:abs}
  f(\vec{x}_i) > f(\vec{x}_j), \quad \forall j \in m
\end{equation}

Similarly, to ensure global optimization capability, $x_j$ here will include individuals in the current generation and in the archived solutions. Further, in updating the archived list, a refining operation is added to limit its size: when a new solution is putting into the archived list, if there is already a solution with a distance less than a predefined $\rho$ in the list, only the more salient solution will be retained among the two. If the salient values of the two solotion are the same, the newer one will be retained.
\subsection{Redistribution Strategy}
Redistribution is to increase the diversity of the population. After a series of iteration, the population tends to converge, and it is not easy to obtain better solutions. A corresponding diversity operation is needed to increase the population's exploration ability further. Due to the existence of the archiving mechanism, it is not necessary for all individuals in the group to converge to the optimal points. Conversely, letting the population explore as many unknown areas as possible will make it desirable to find more solutions. Nonetheless, to improve the accuracy of the algorithm, we also hope to find absolutely salient solutions around relatively salient ones. Therefore, the redistribution strategy here does not redistribute all individuals but only individuals in the non-salient list. Herein, the redistribution operation will be determined according to the archive change, i.e., after a continuous $t'$ iteration steps, the archived list remained unchanged, the non-salient individuals will be redistributed.


    





\section{Results}
We tested the proposed method on some representative benchmark functions. The function introduction, parameter configuration, and some preliminary results will be presented in this section.
\subsection{Test Functions and Configurations}
This paper chooses some 1D and 2D functions from CEC2013 benchmark for visually testing \cite{li2013benchmark}. This benchmark is a representative one in the field of multimodal optimization. It contains a total of 20 functions, all of which are formulated as maximization problems. For the convenience of visualization, we selected some 1D and 2D functions to verify the proposed method. The basic properties of the chosen functions are shown in Table \ref{tab:bf}
. The MaxFes is maximum evaluation times for each function, the PH is the Peak Height, and the $\rho$ is the niching radius that can distinguish two closest global optima. The population size of tests are set to 100, and the maximum iteration steps are set to Maxfes/pop. Other algorithm parameter configurations are shown in Table \ref{tab:config}.
\begin{table}[!htb]
  \caption{Benchmark Functions}\label{tab:bf}
  \centering
\begin{tabular}{@{}clcccc@{}}
\toprule
No. & \multicolumn{1}{c}{Function Name}  & MaxFes & PH        & $\rho$ \\ \midrule
F1   & Five-Uneven-Peak Trap     (1D)  & 5E4   & 200.0     & 0.01   \\
F2   & Equal Maxima              (1D) & 5E4   & 1.0       & 0.01   \\
F3   & Uneven Decreasing Maxima  (1D)  & 5E4   & 1.0       & 0.01   \\
F4   & Himmelblau                (2D)  & 5E4   & 200.0     & 0.01   \\
F5   & Six-Hump Camel Back       (2D)  & 5E4   & 1.03163   & 0.5    \\
F6   & Shubert                   (2D)  & 2E5  & 186.731   & 0.5    \\
F7  & Modified Rastrigin        (2D)  & 2E5  & -2.0      & 0.01   \\
  \bottomrule
\end{tabular}
\end{table}


\begin{table}[!htb]
   \caption{Parameter Configurations}\label{tab:config}
  \centering
\begin{tabular}{@{}ccccc|ccc@{}}
\toprule
\multicolumn{5}{c}{BSO Parameters}             & \multicolumn{3}{|c}{ABSO Parameters} \\ \midrule
$perc\%$ & $p_{d}$ & $p_{one}$ & $p_{e}$ & $k$ & $m$      & $\sigma^2$     & $t'$     \\ \midrule
0.1      & 0.1     & 0.8       & 0.8     & 25  & 20       & 0.4            & 5        \\ \bottomrule
\end{tabular}
\end{table}

\subsection{Results}
\begin{figure*}[!htb]
  \centering
  \subfigure[F1 Results]{\label{fig:f1} \includegraphics[trim=50 210 60 210,clip,width=0.32\textwidth]{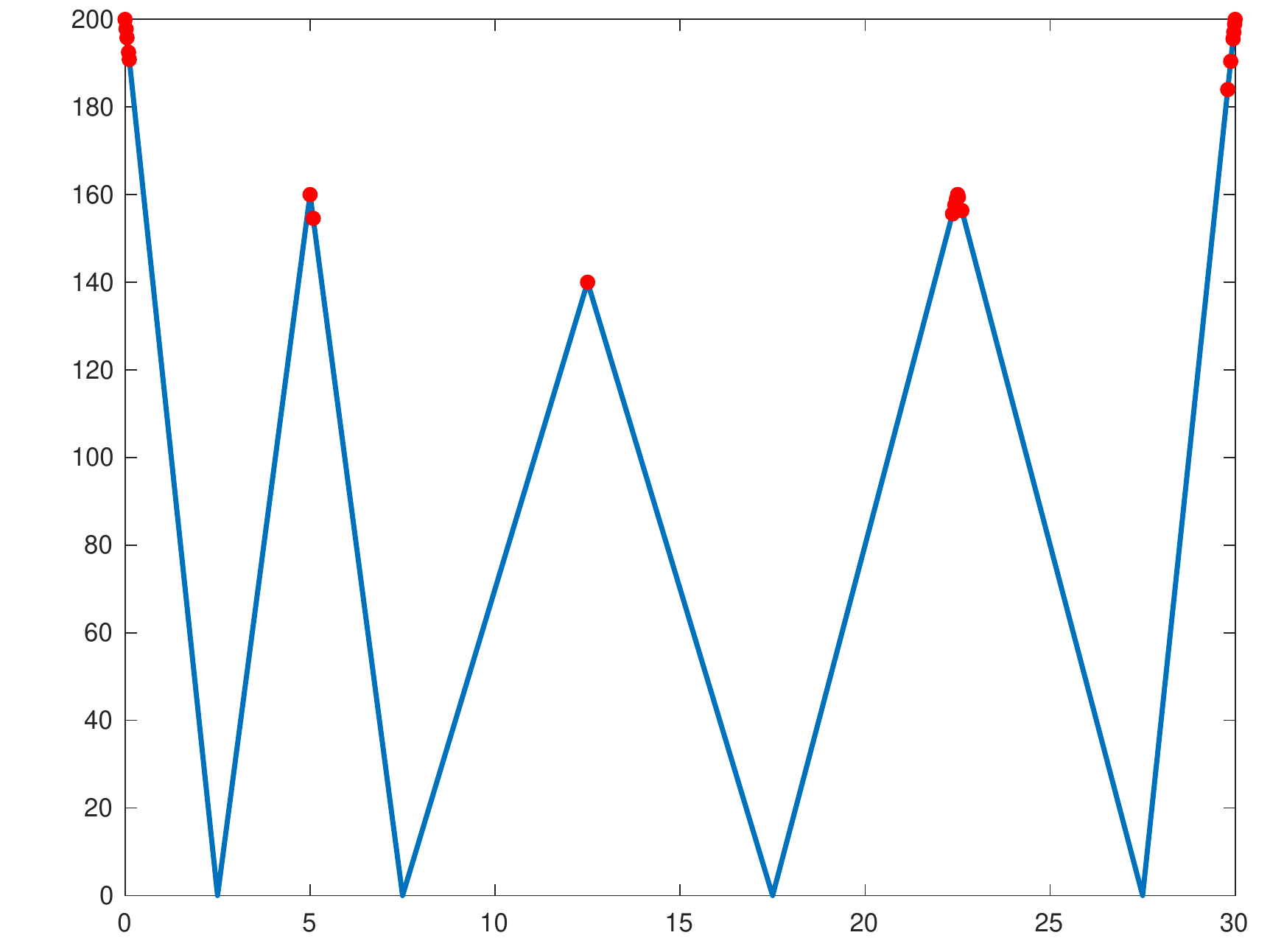}}
  \subfigure[F1 Fitness Convergence]{\label{fig:f1rf} \includegraphics[trim=50 210 60 210,clip,width=0.32\textwidth]{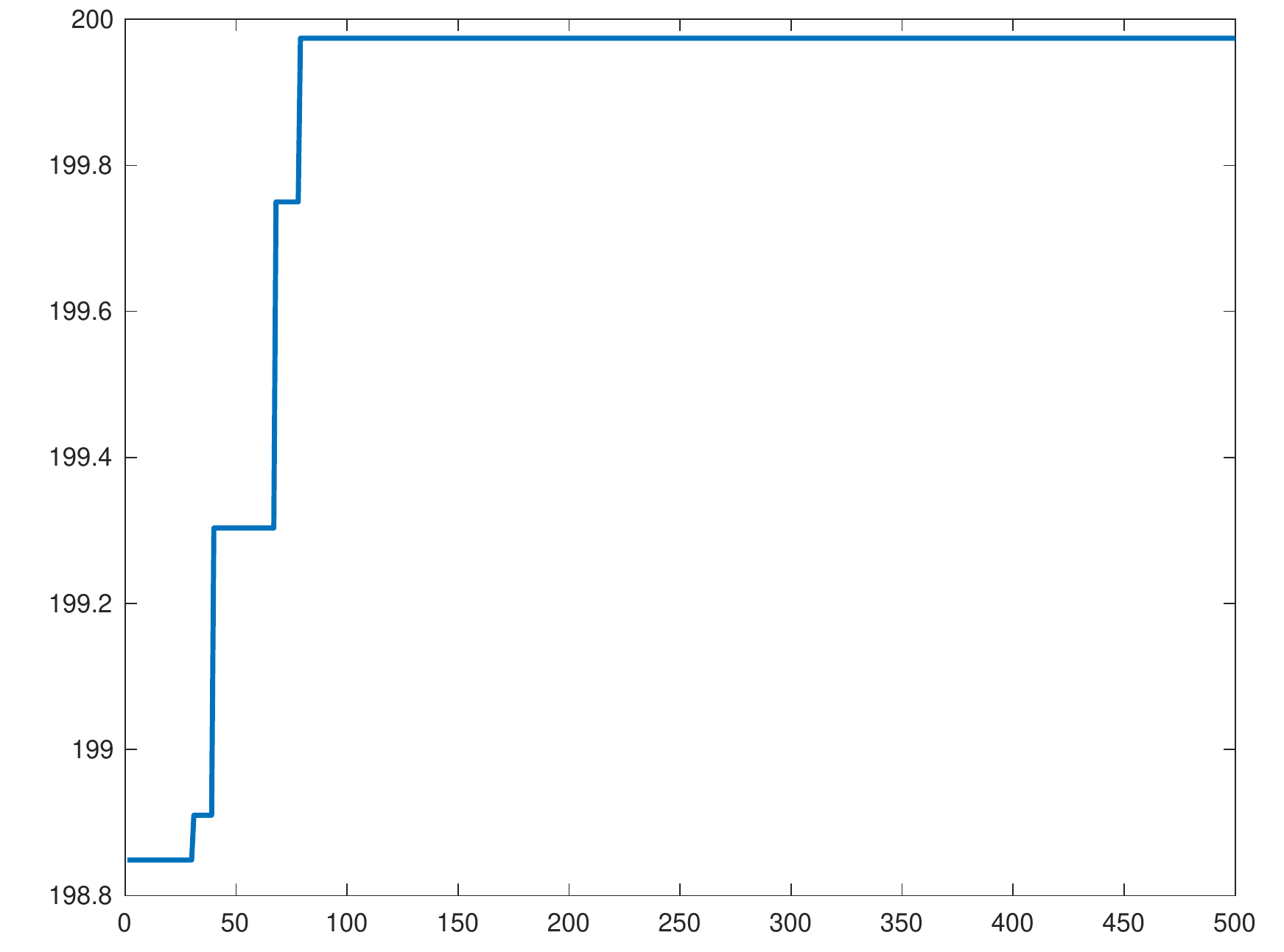}}
  \subfigure[F1 Attention Convergence]{\label{fig:f1ra} \includegraphics[trim=50 210 60 210,clip,width=0.32\textwidth]{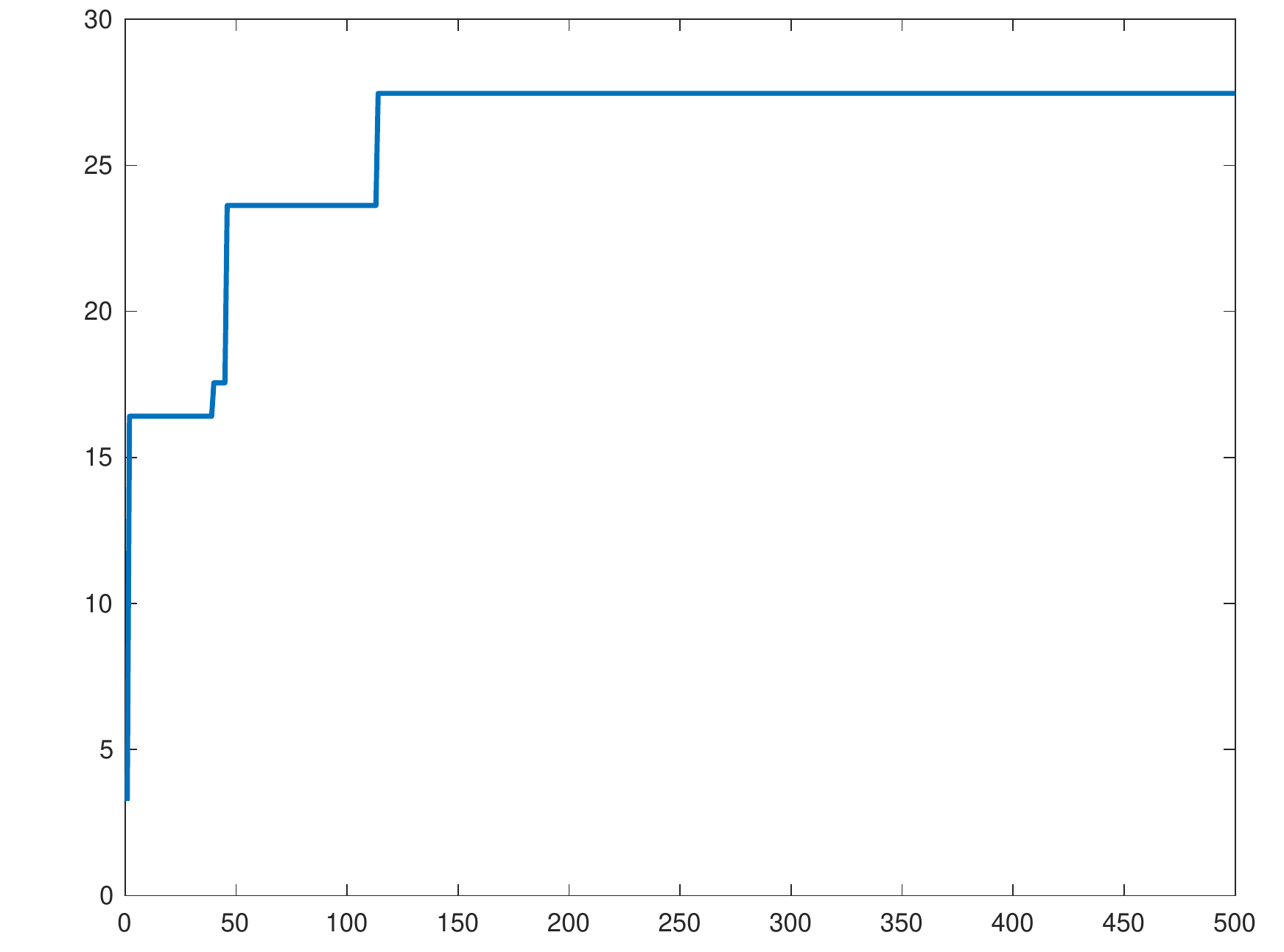}}

  \subfigure[F2 Results]{\label{fig:f2} \includegraphics[trim=50 210 60 210,clip,width=0.32\textwidth]{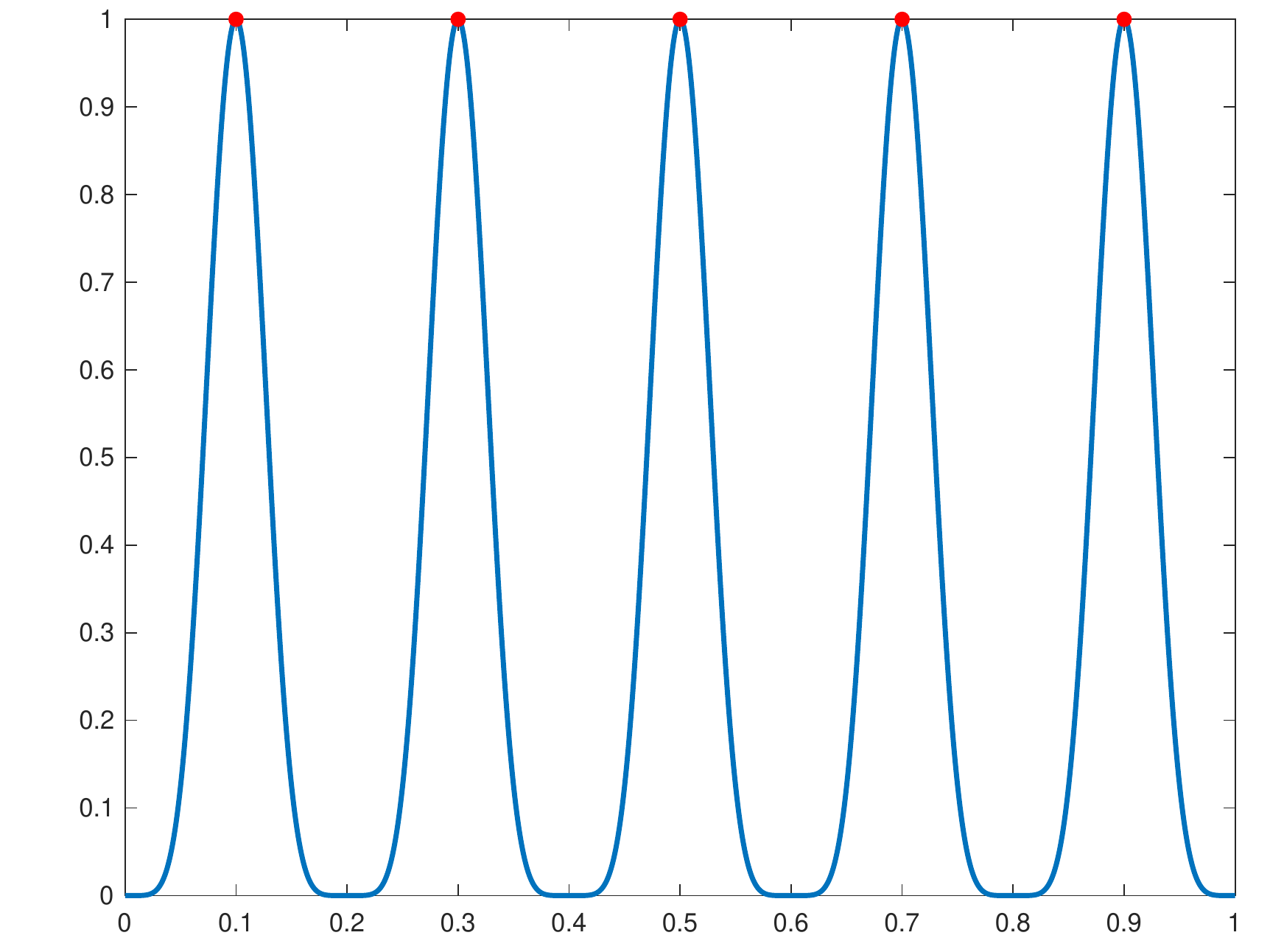}}
  \subfigure[F2 Fitness Convergence]{\label{fig:f2rf} \includegraphics[trim=50 210 60 210,clip,width=0.32\textwidth]{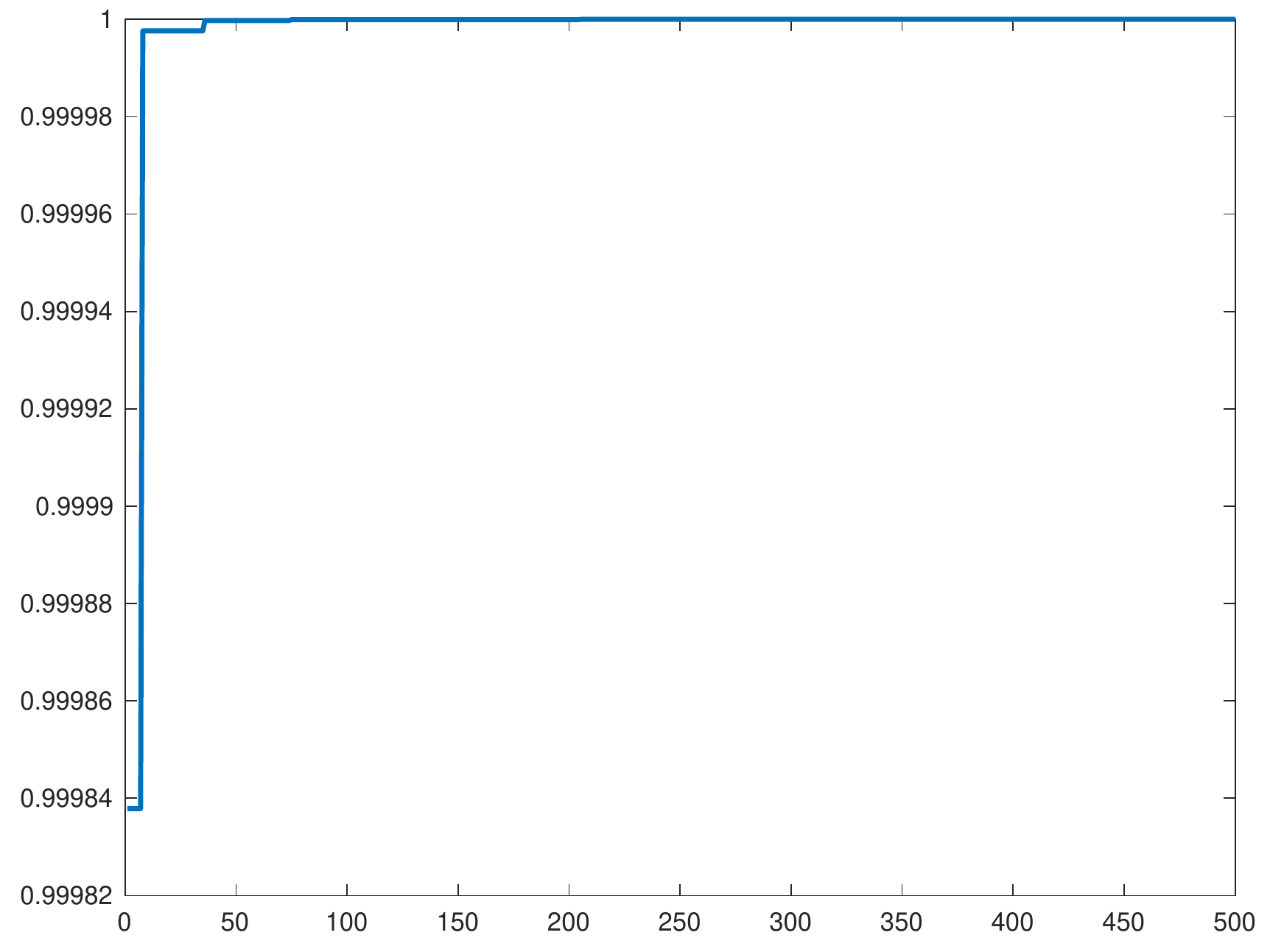}}
  \subfigure[F2 Attention Convergence]{\label{fig:f2ra} \includegraphics[trim=50 210 60 210,clip,width=0.32\textwidth]{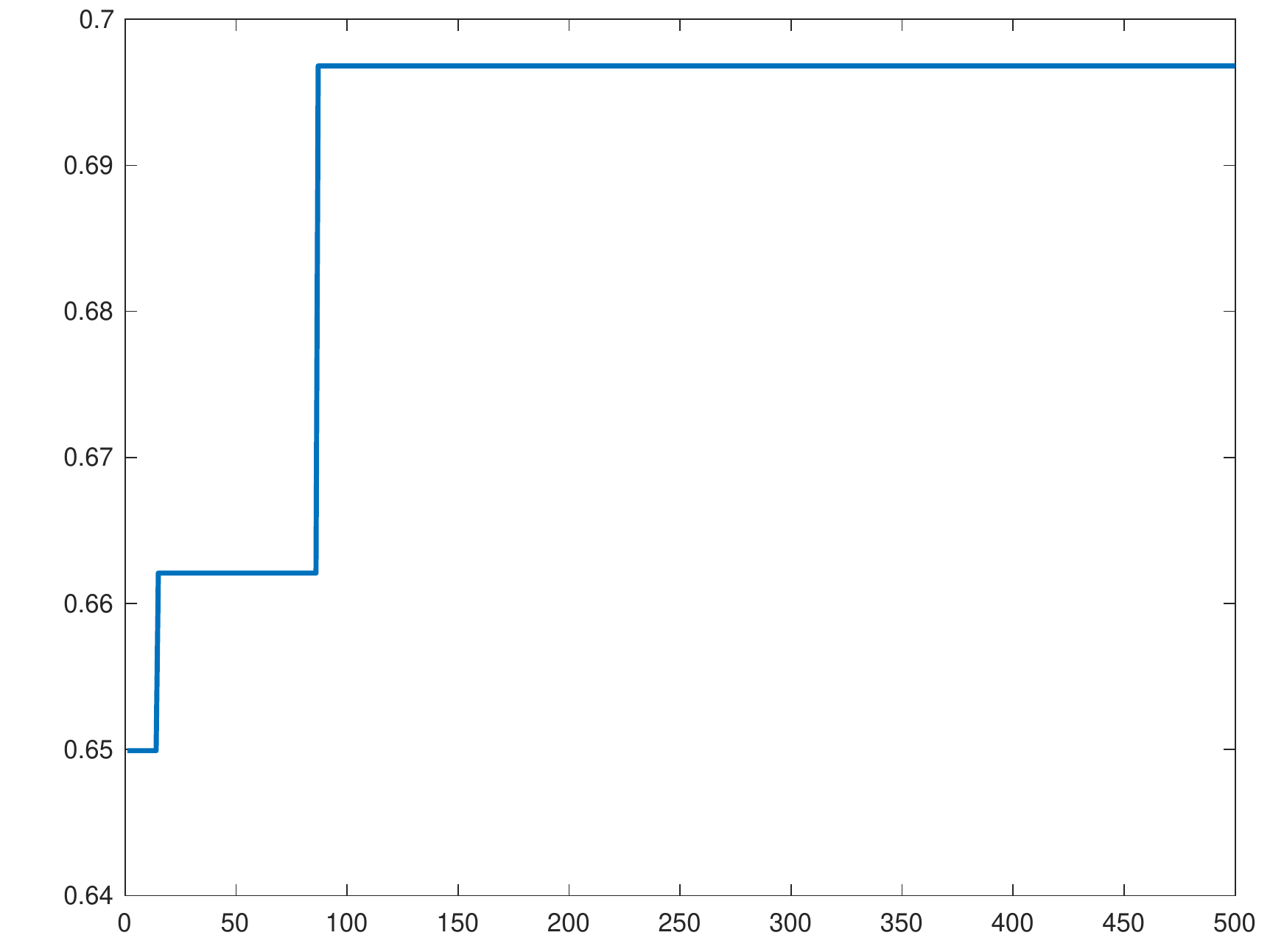}}
  
  \subfigure[F3 Results]{\label{fig:f3} \includegraphics[trim=50 210 60 210,clip,width=0.32\textwidth]{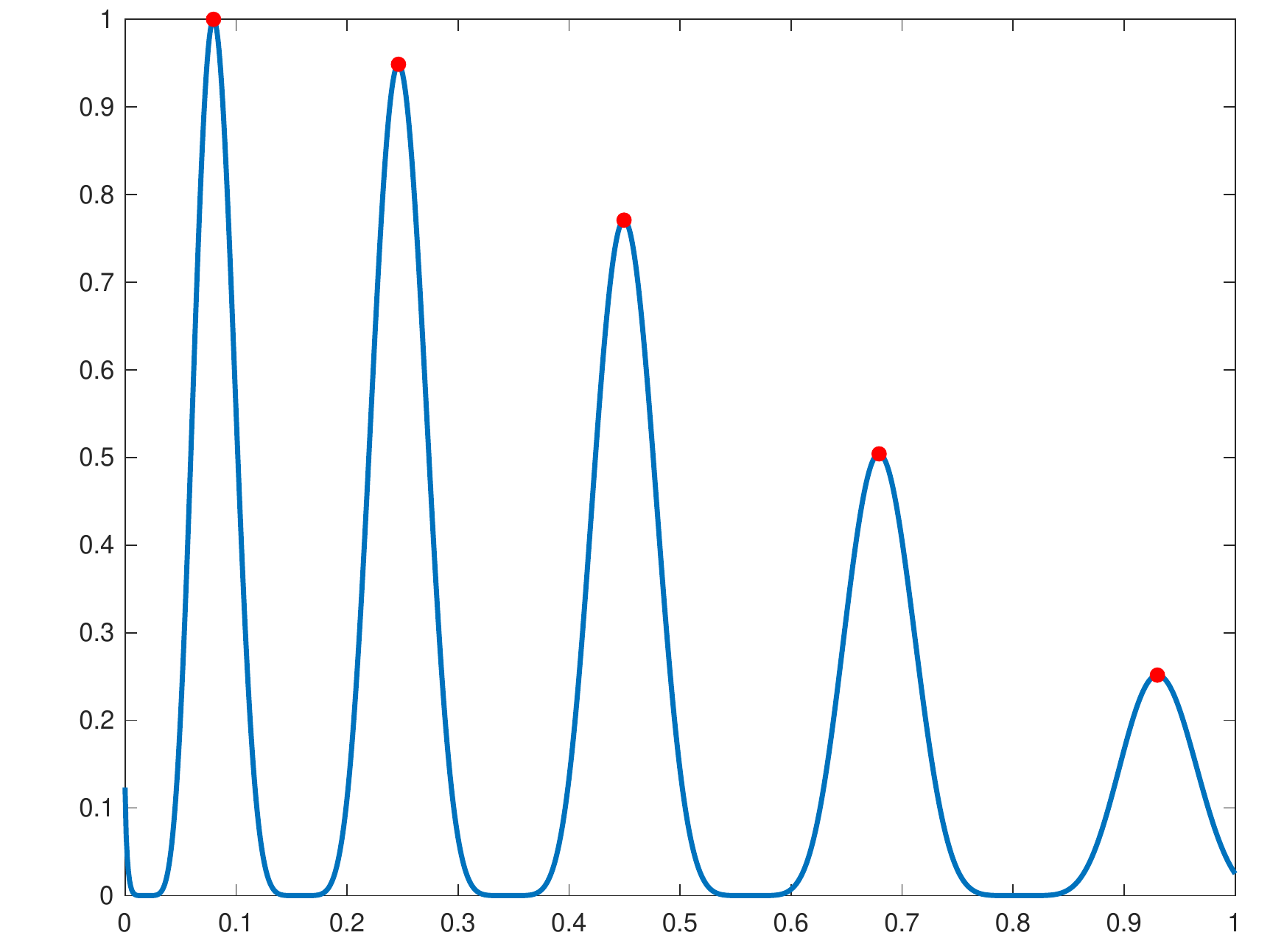}}
  \subfigure[F3 Fitness Convergence]{\label{fig:f3rf} \includegraphics[trim=50 210 60 210,clip,width=0.32\textwidth]{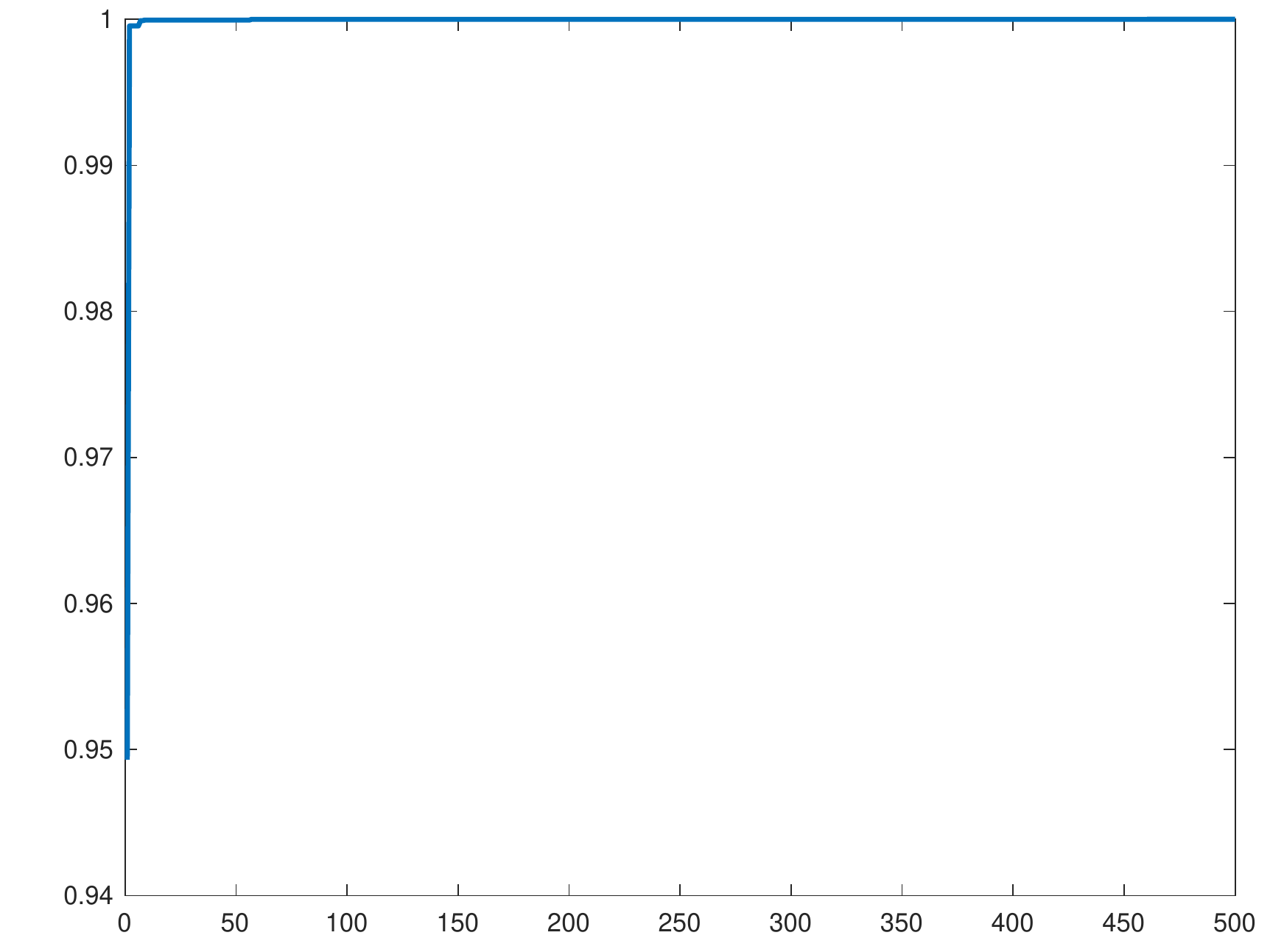}}
  \subfigure[F3 Attention Convergence]{\label{fig:f3ra} \includegraphics[trim=50 210 60 210,clip,width=0.32\textwidth]{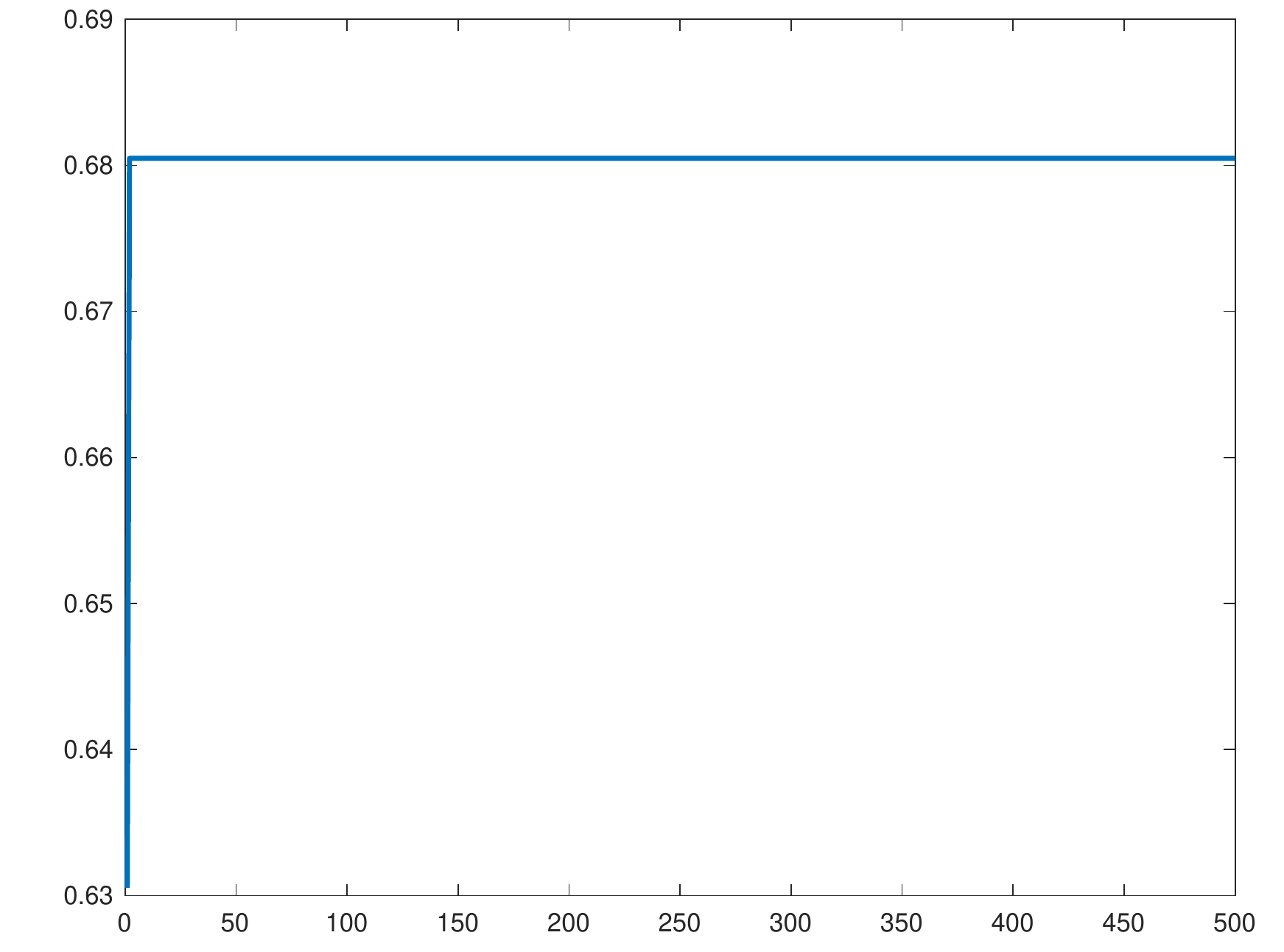}}
  \caption{Results for 1D Functions}
  \label{fig:visual}
\end{figure*}

One set of the test results of 1D multi-modal functions are shown in Fig.\ref{fig:visual}. In the figure, each row from left to right are solutions, fitness value convergence process, and attention value convergence process, respectively.

It can be seen that both the global maximum and the local maximum can be successfully found. For example, there are two global maximums and three local maximums in Fig.\ref{fig:f1}, one global optimal with three local peaks in Fig.\ref{fig:f3}, and all are global maximums in Fig.\ref{fig:f2}.

Besides, from Fig.\ref{fig:f1rf}, \ref{fig:f2rf}, and \ref{fig:f3rf}, we know the convergence speed in the test results is relatively fast, and they all reach the global optimum within a short time after the start of the iteration. The maximum number of iterations of the test for the above three functions is 500 generations. We have tested each function 30 times. The average fitness convergence generations of F1, F2, and F3 are 155, 87, and 17, respectively. This result shows that the proposed attention-oriented method can guide the search quickly converge to the global optimal, which has a high search efficiency.

Furthermore, we also recorded the changes of maximum saliency values in each generation during the iteration, shown in Figures 1, 2, and 3, respectively. It can be seen that in the iterative process, the most salient value in each generation also shows an upward trend. It is worth noting that after the fitness value has converged to the global optimum, the saliency value still changes. This because neighbors used by each individual to calculate its saliency value are dynamic changing. Such a saliency value change shows that the algorithm tries to guide the population to find more salient solutions, which may be another peak.

\begin{figure*}[!htb]
    \centering
    \subfigure[F4]{\label{fig:f4} \includegraphics[trim=50 210 60 210,clip,width=0.23\textwidth]{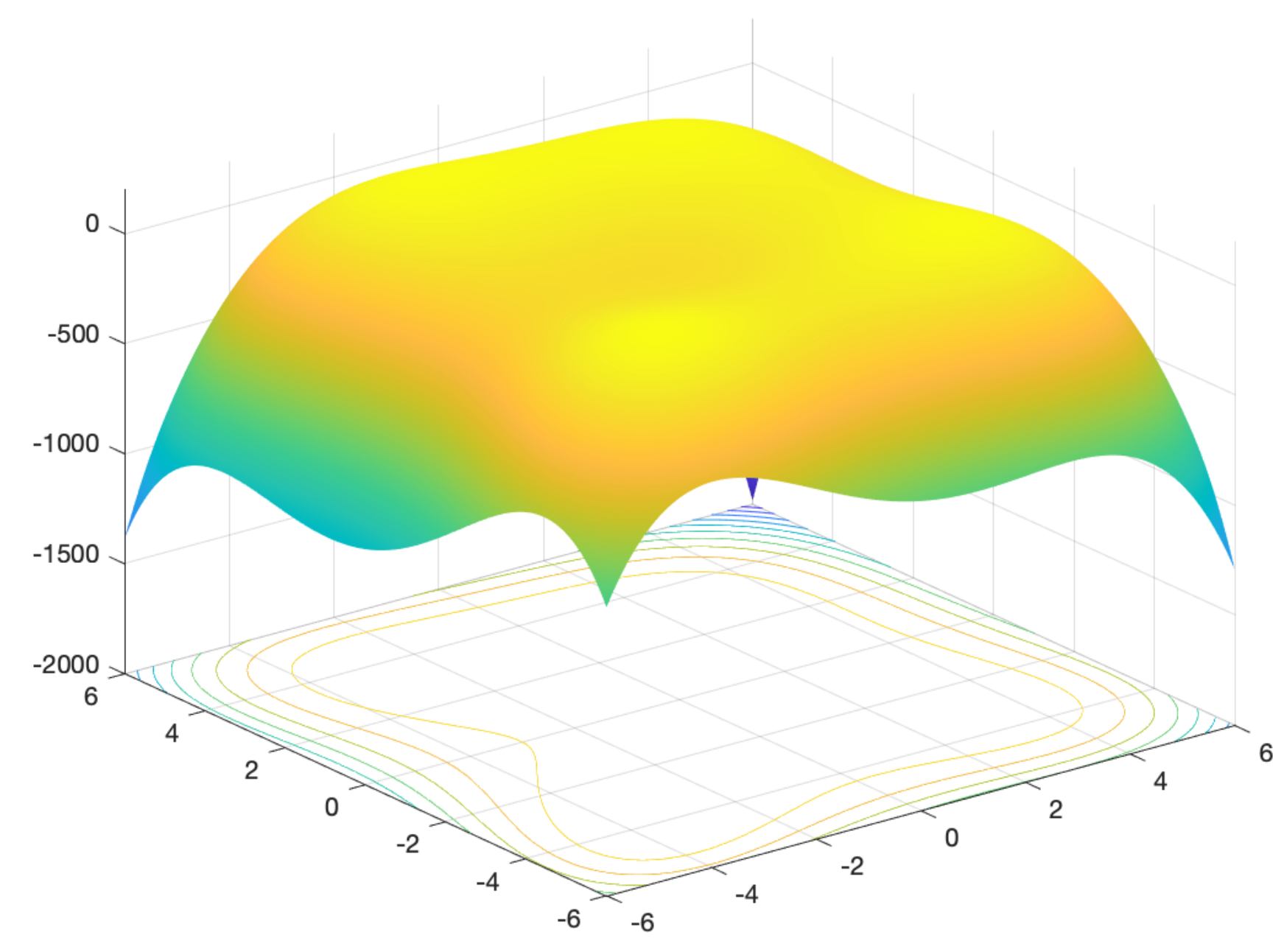}}
    \subfigure[F4 Results]{\label{fig:f4r} \includegraphics[trim=50 210 60 210,clip,width=0.23\textwidth]{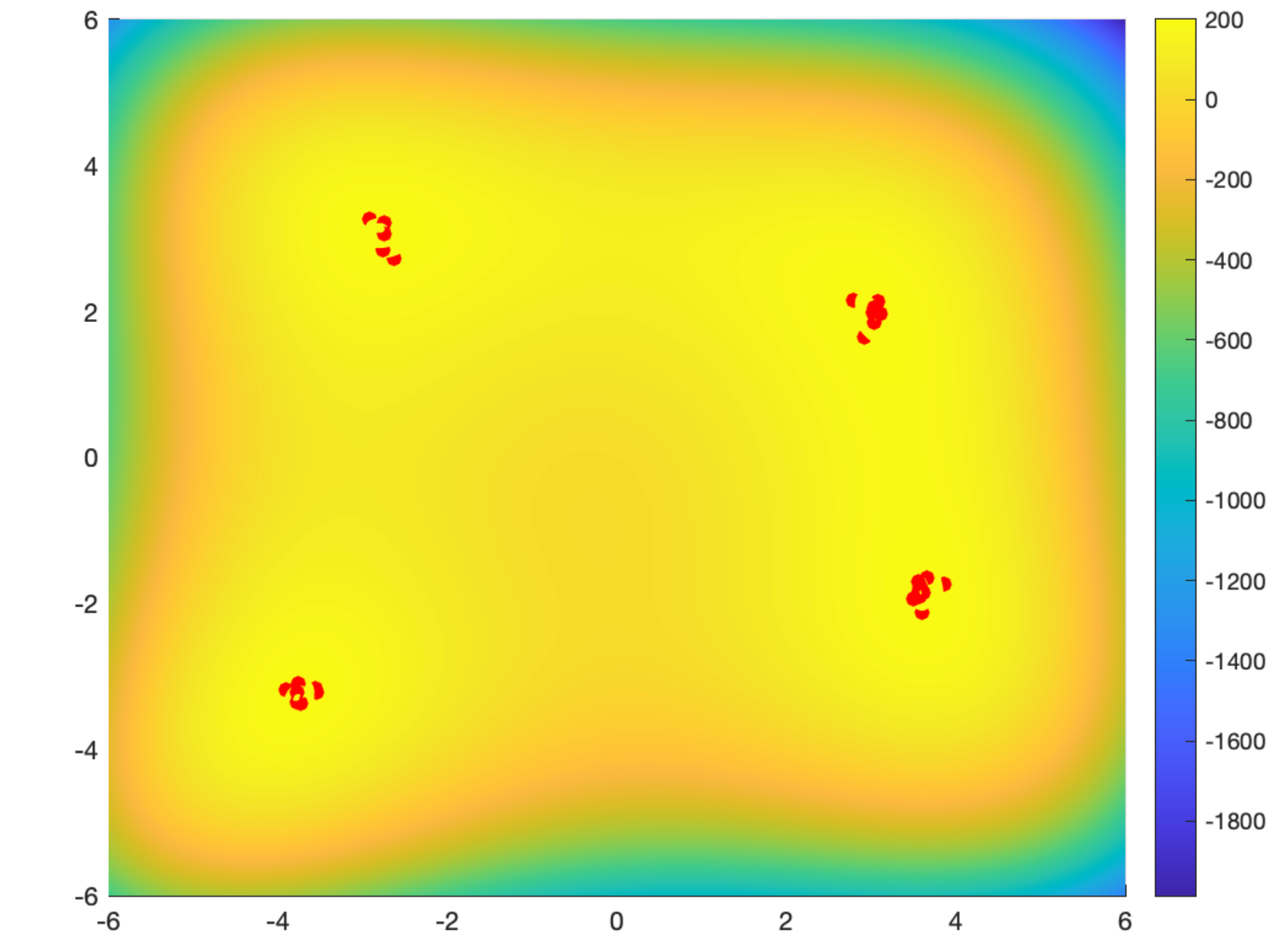}}
    \subfigure[F4 Fitness Convergence]{\label{fig:f4f} \includegraphics[trim=50 210 60 210,clip,width=0.23\textwidth]{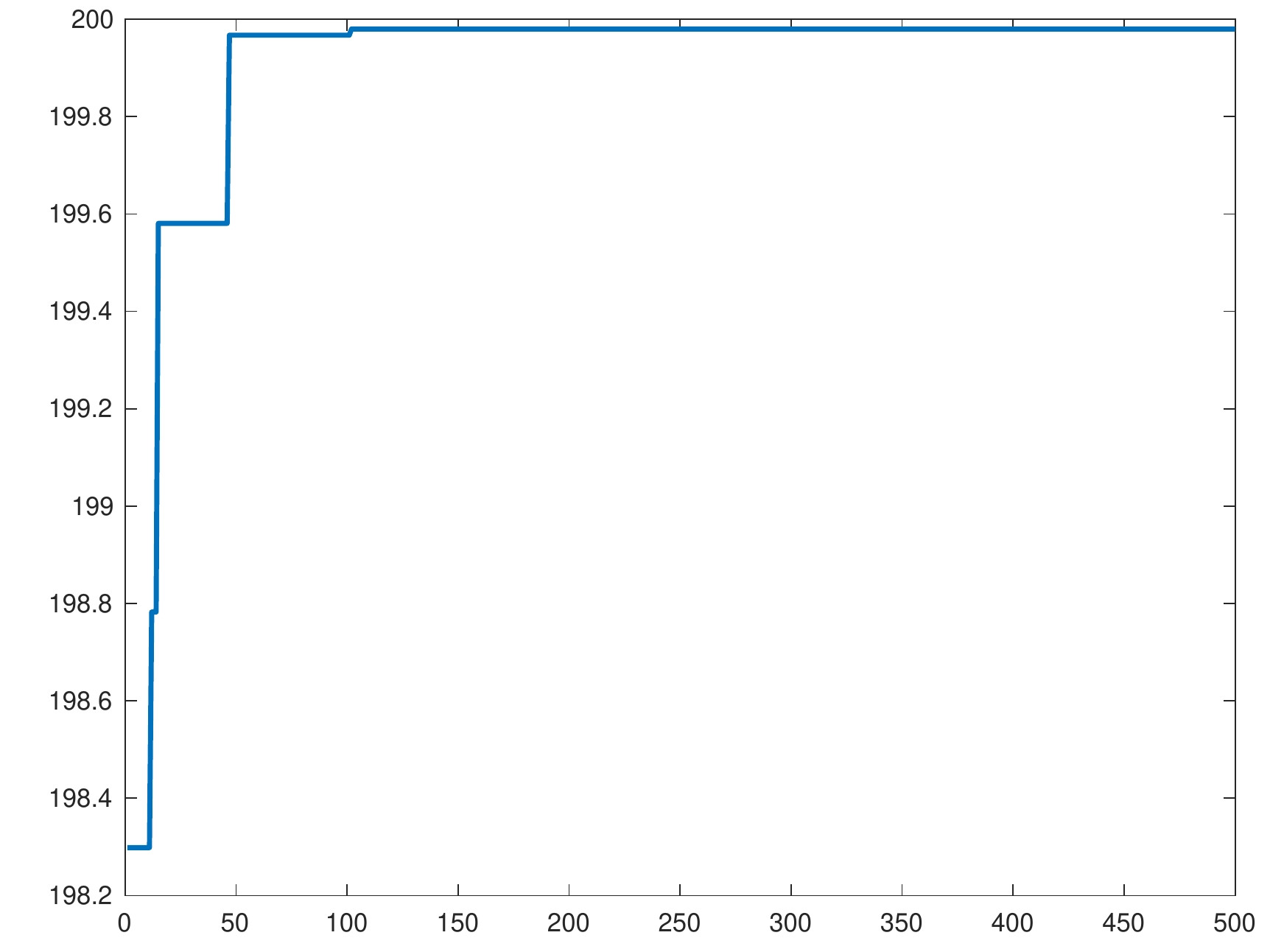}}
    \subfigure[F4 Attention Convergence]{\label{fig:f4a} \includegraphics[trim=50 210 60 210,clip,width=0.23\textwidth]{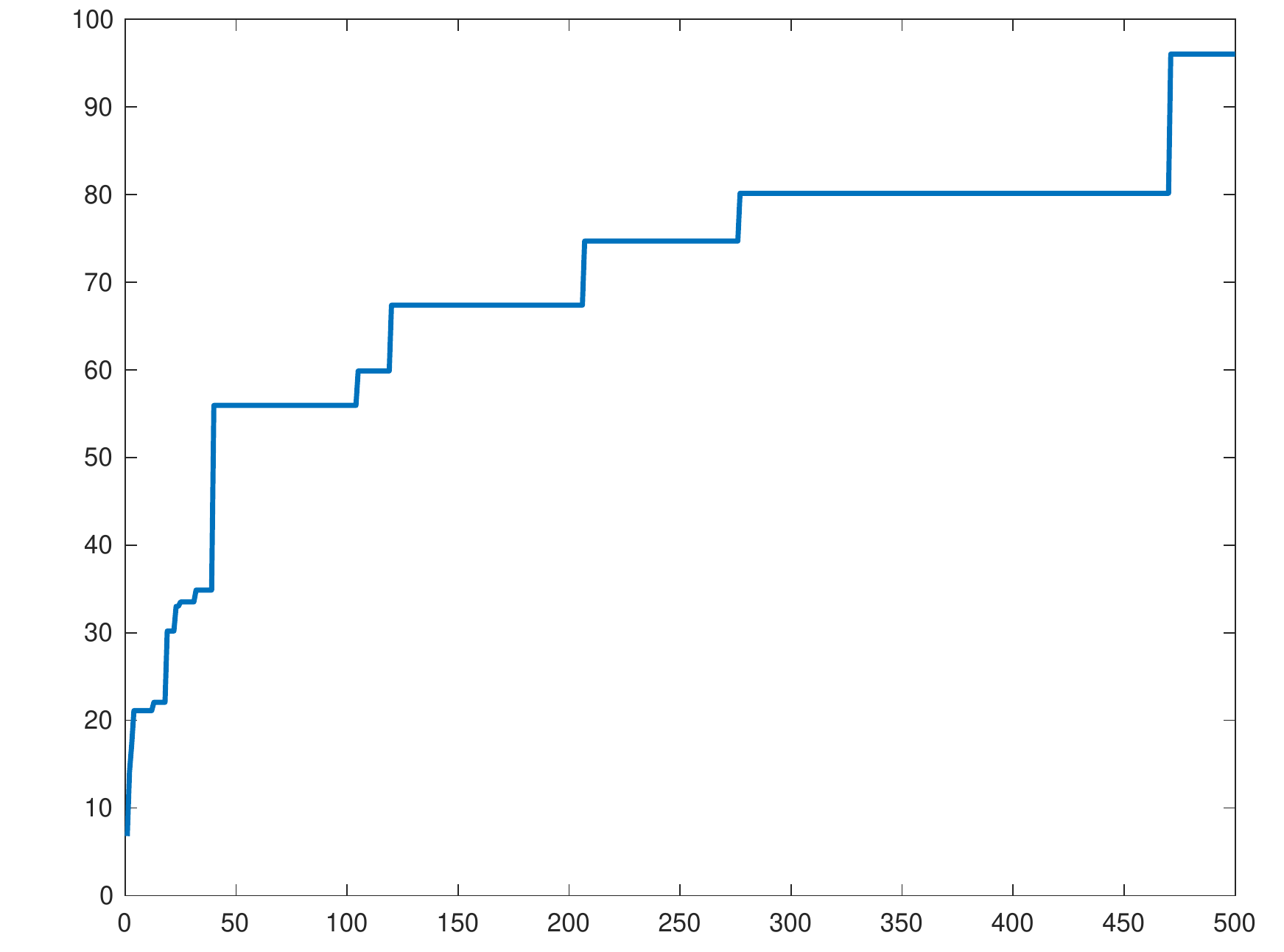}}
 
    \subfigure[F5]{\label{fig:f5} \includegraphics[trim=50 210 60 210,clip,width=0.23\textwidth]{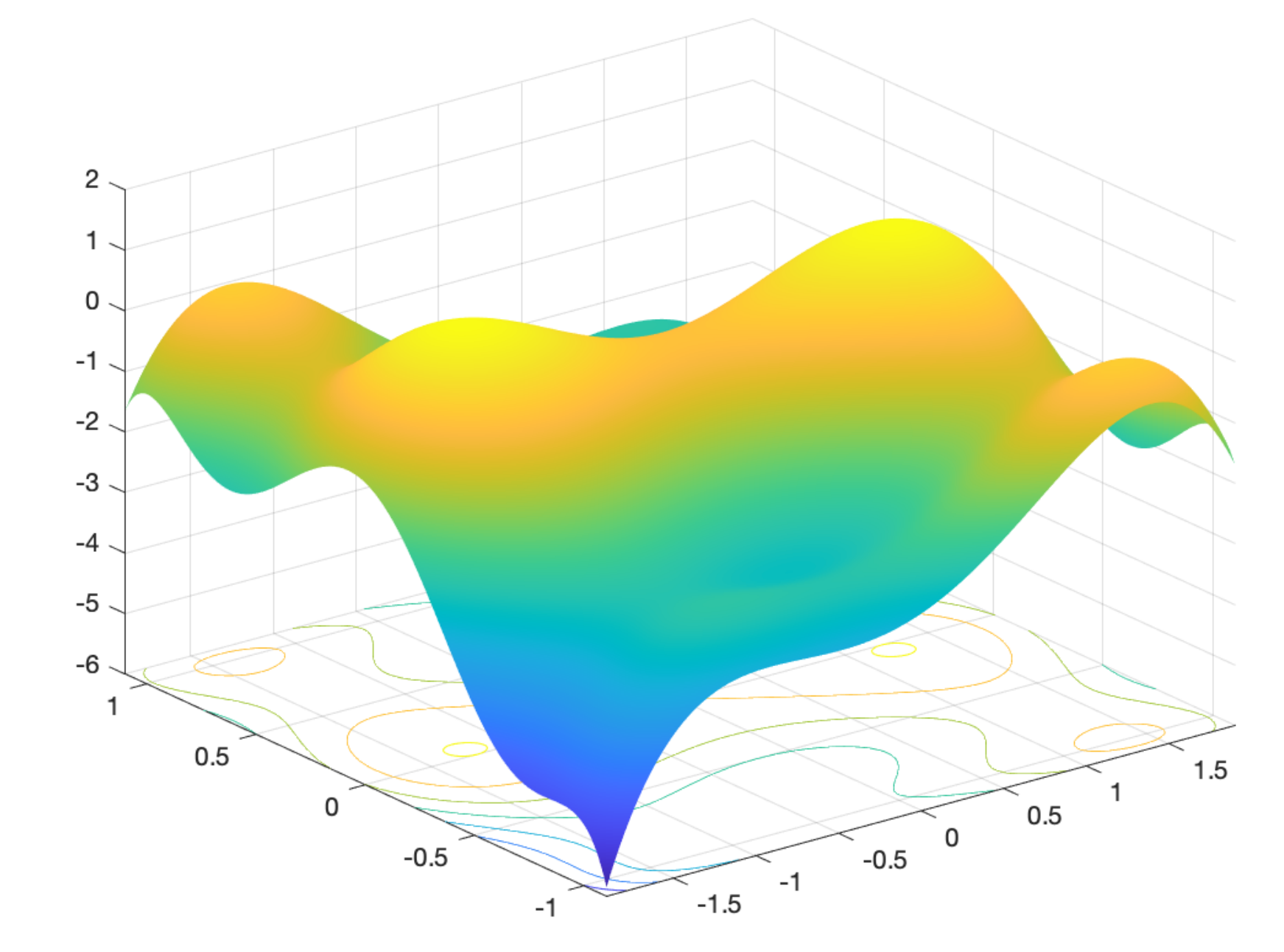}}
    \subfigure[F5 Results]{\label{fig:f5r} \includegraphics[trim=50 210 60 210,clip,width=0.23\textwidth]{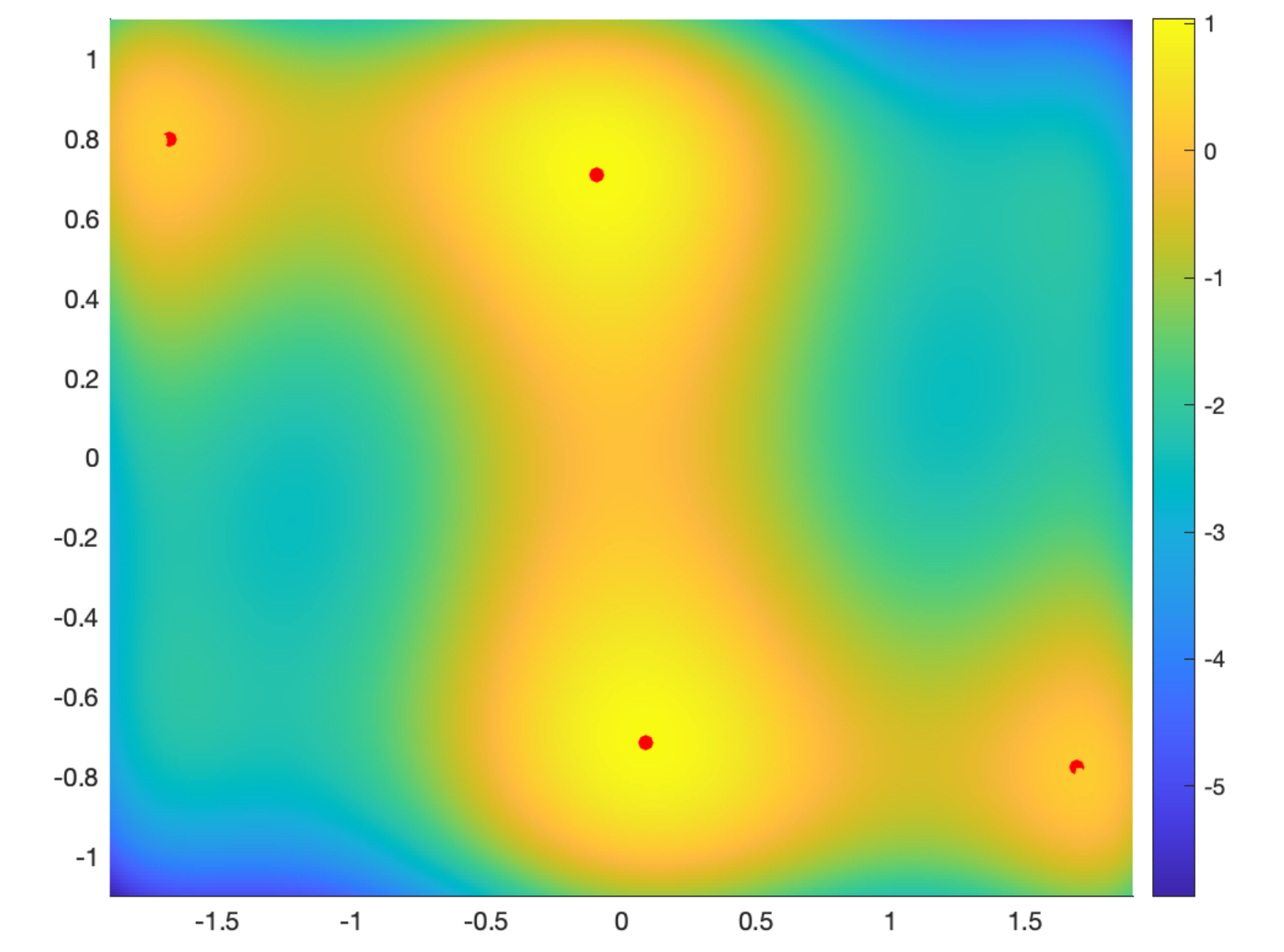}}
    \subfigure[F5 Fitness Convergence]{\label{fig:f5f} \includegraphics[trim=50 210 60 210,clip,width=0.23\textwidth]{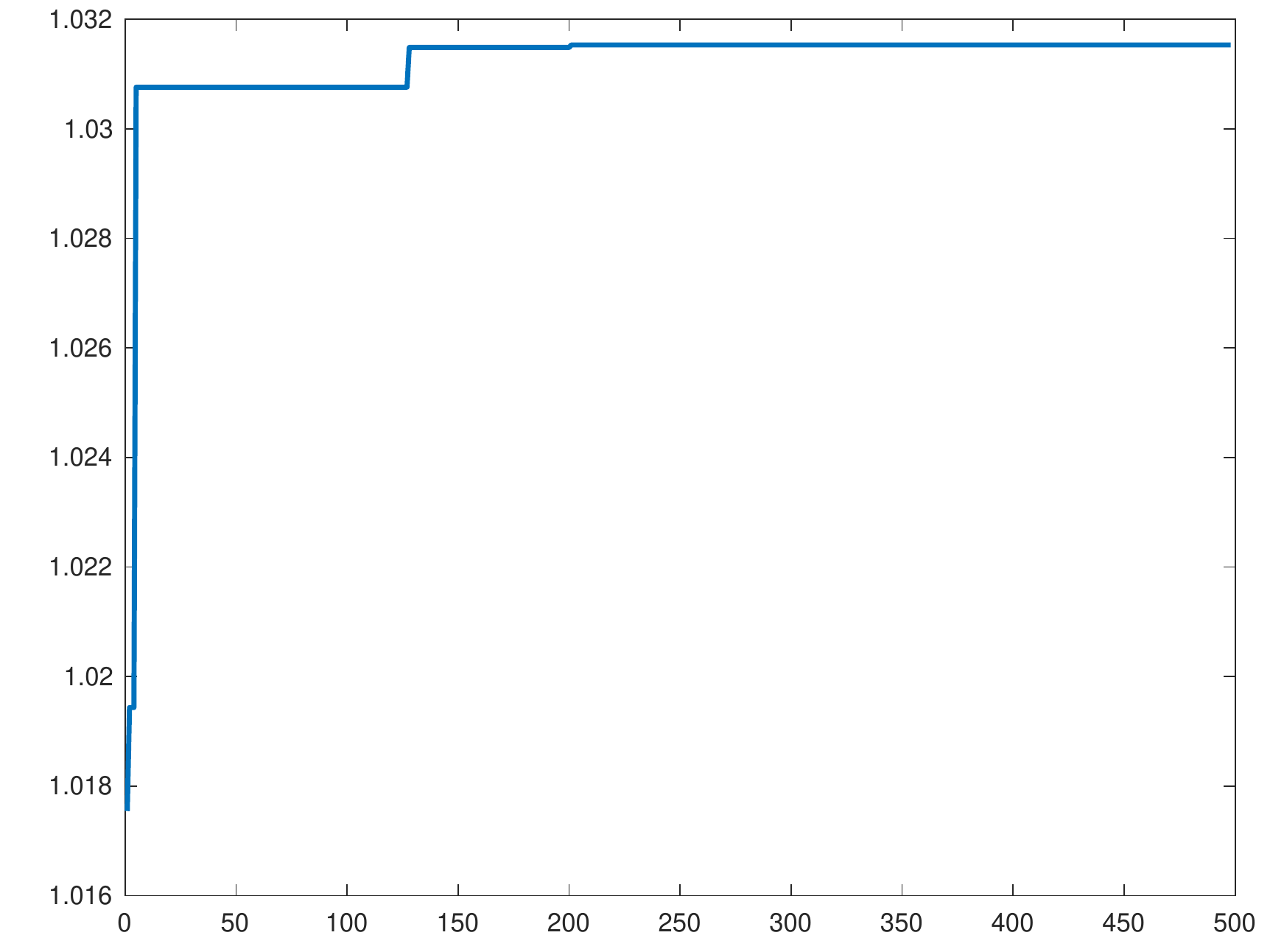}}
    \subfigure[F5 Attention Convergence]{\label{fig:f5a} \includegraphics[trim=50 210 60 210,clip,width=0.23\textwidth]{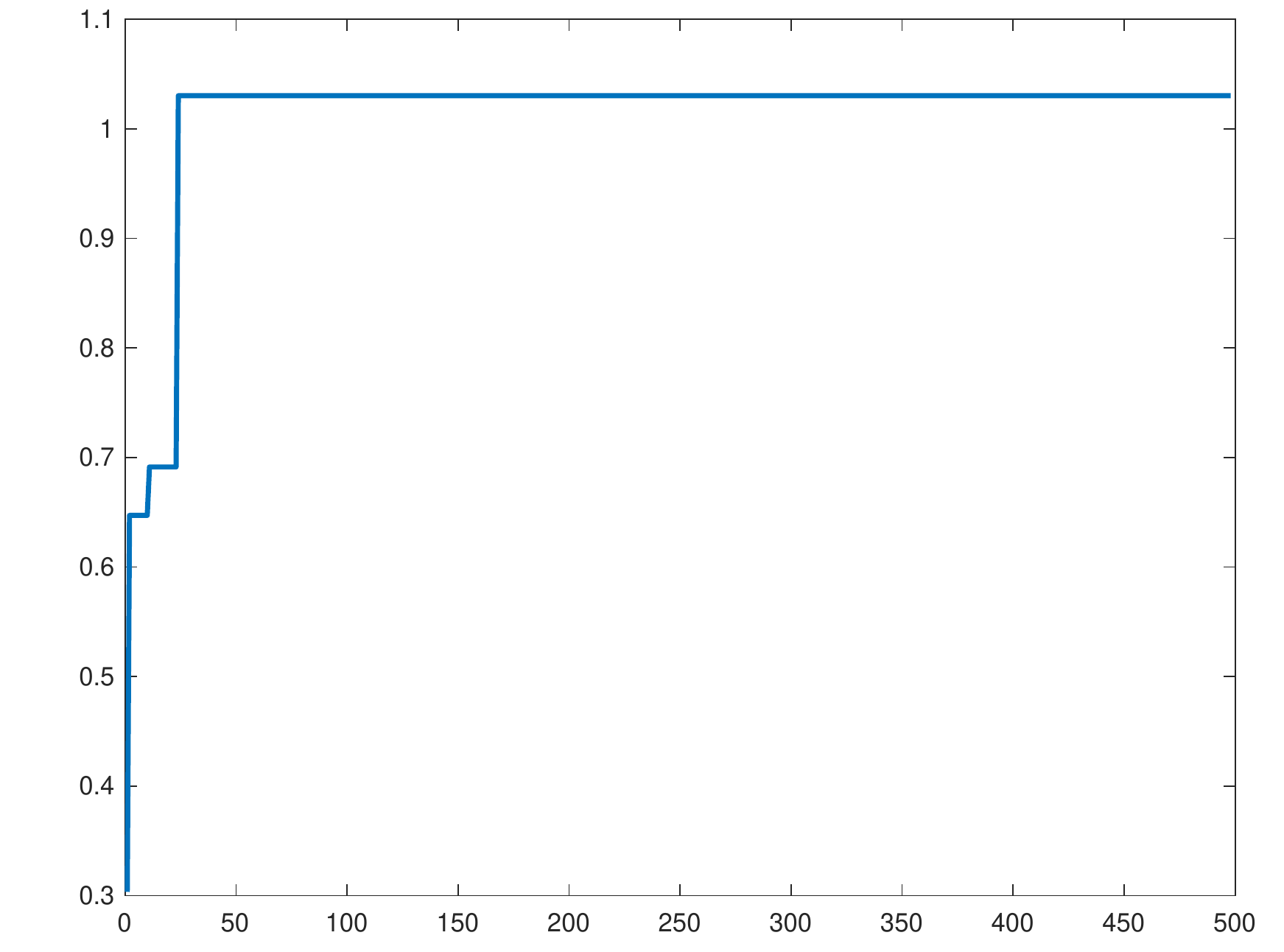}}

    \subfigure[F6]{\label{fig:f6} \includegraphics[trim=50 210 55 210,clip,width=0.23\textwidth]{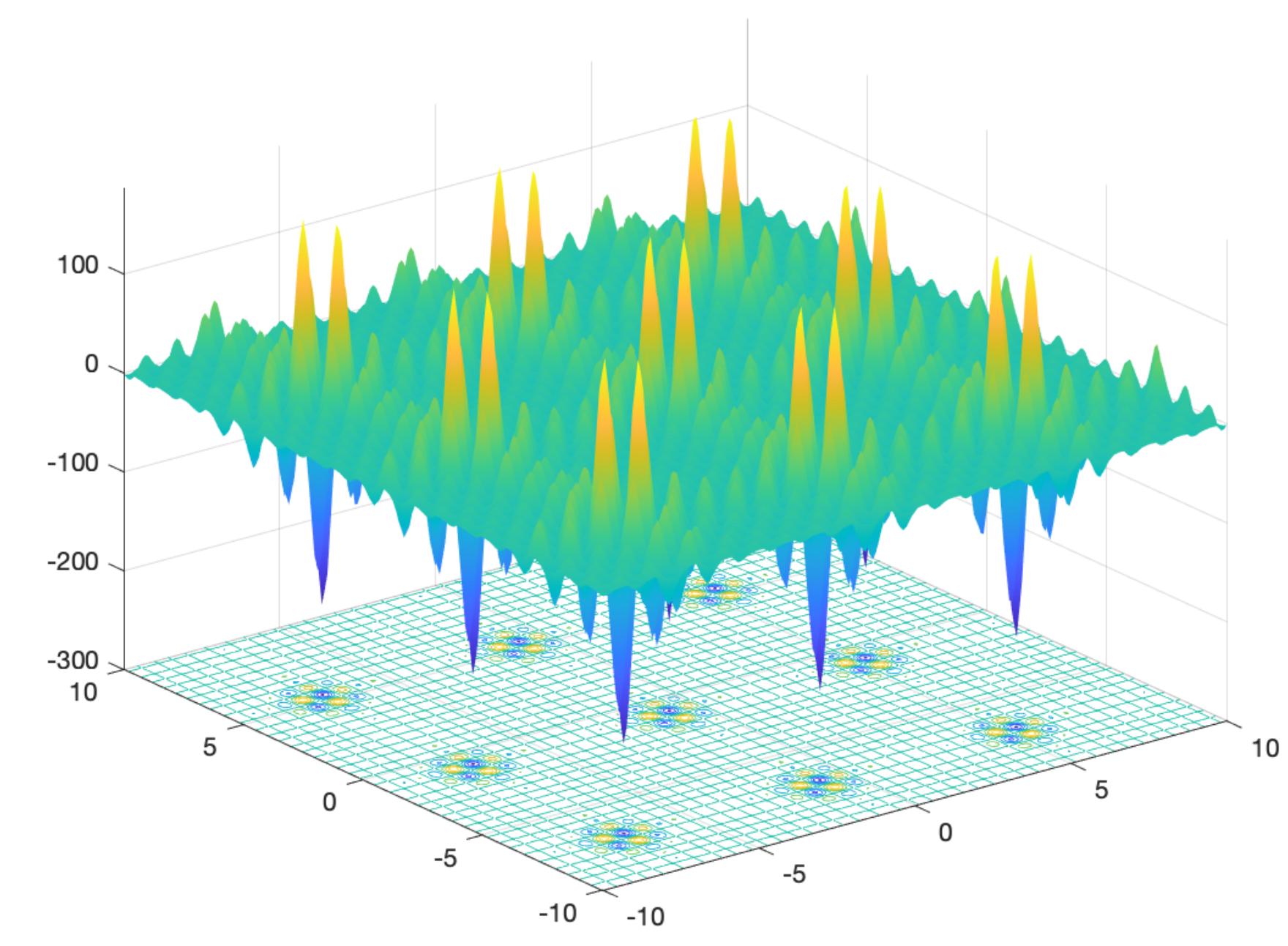}}
    \subfigure[F6 Results]{\label{fig:f6r} \includegraphics[trim=50 210 60 210,clip,width=0.23\textwidth]{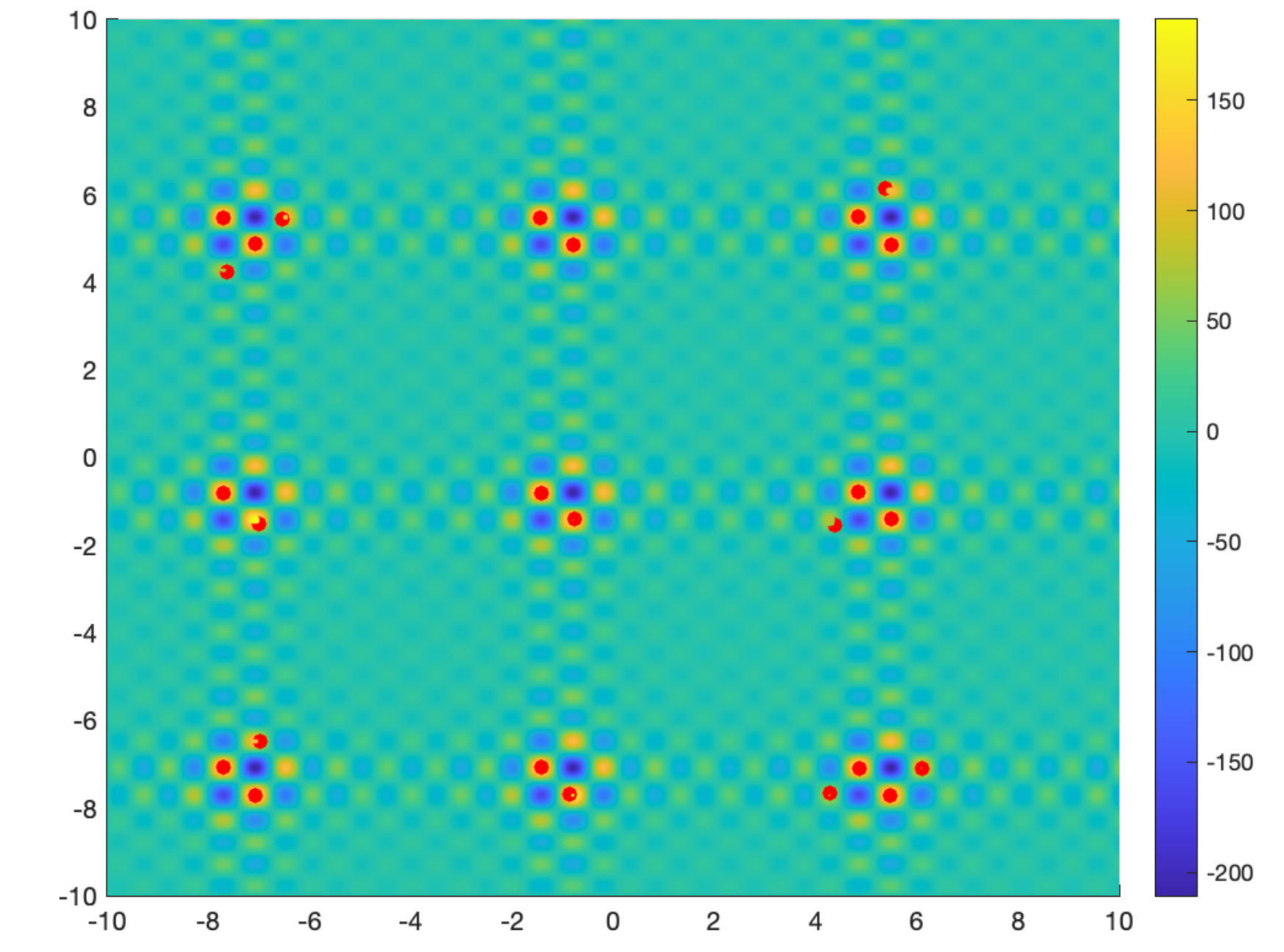}}
    \subfigure[F6 Fitness Convergence]{\label{fig:f6f} \includegraphics[trim=50 210 60 210,clip,width=0.23\textwidth]{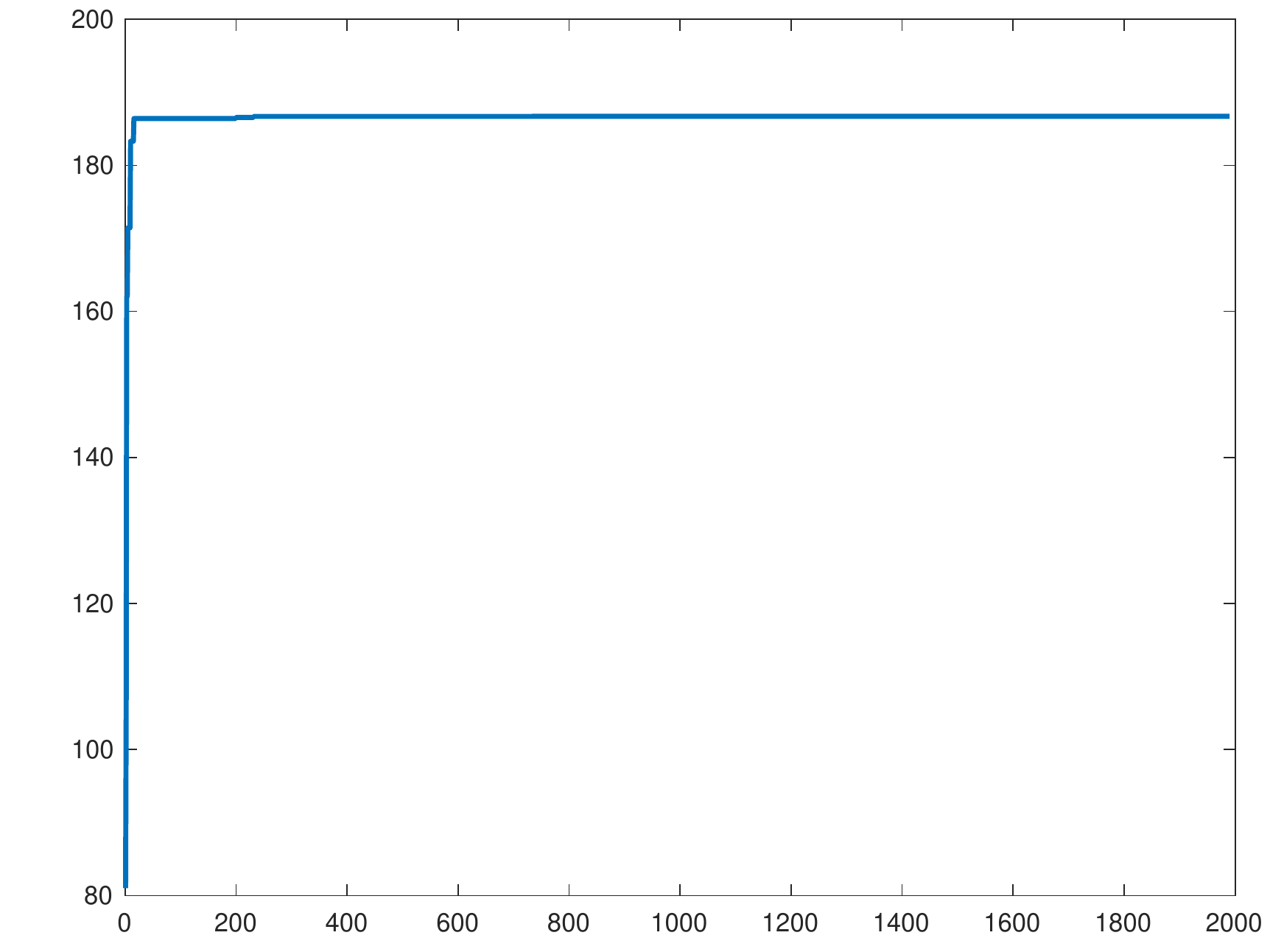}}
    \subfigure[F6 Attention Convergence]{\label{fig:f6a} \includegraphics[trim=50 210 60 210,clip,width=0.23\textwidth]{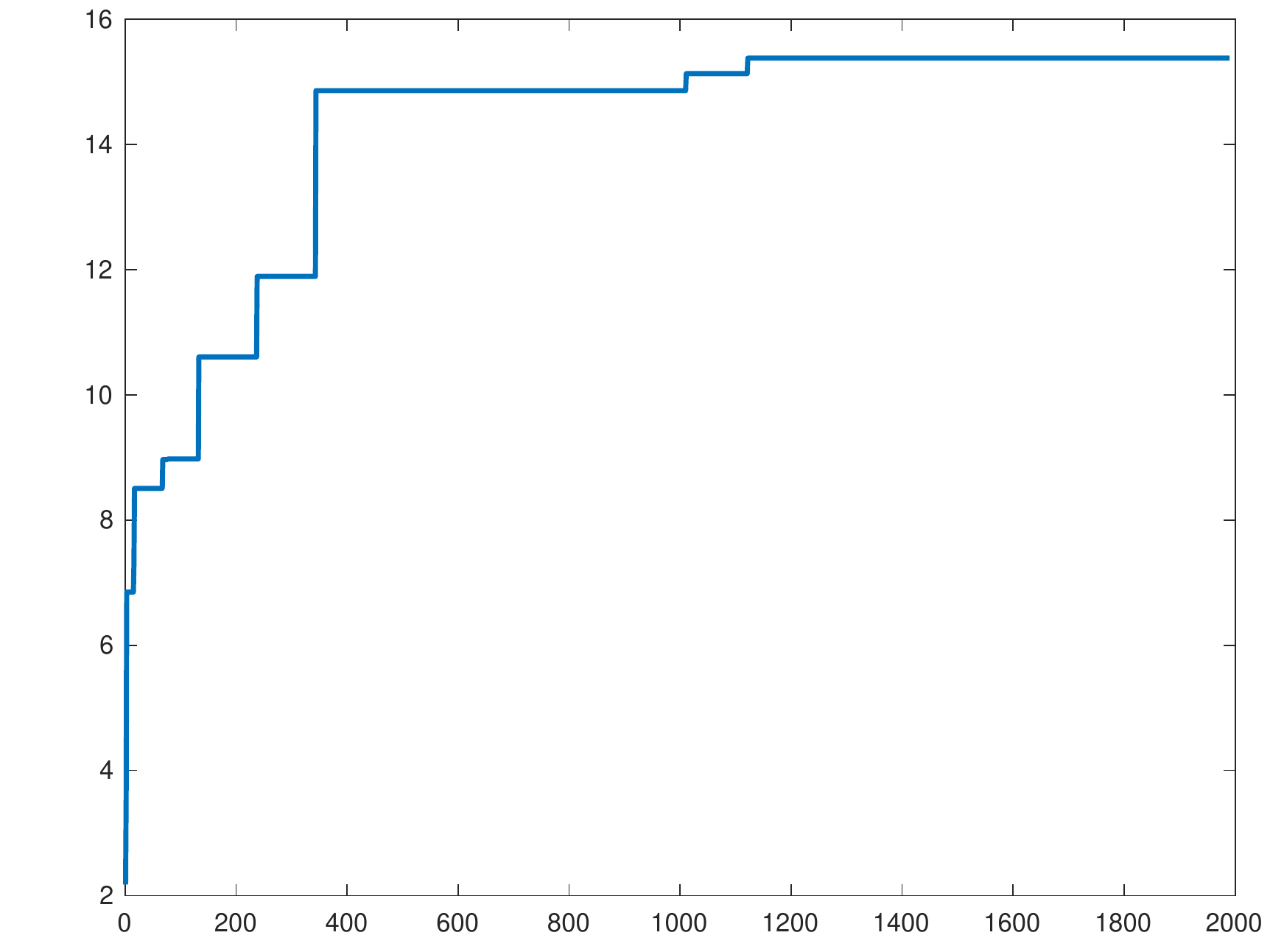}}
 
    \subfigure[F7]{\label{fig:f7} \includegraphics[trim=50 210 60 210,clip,width=0.23\textwidth]{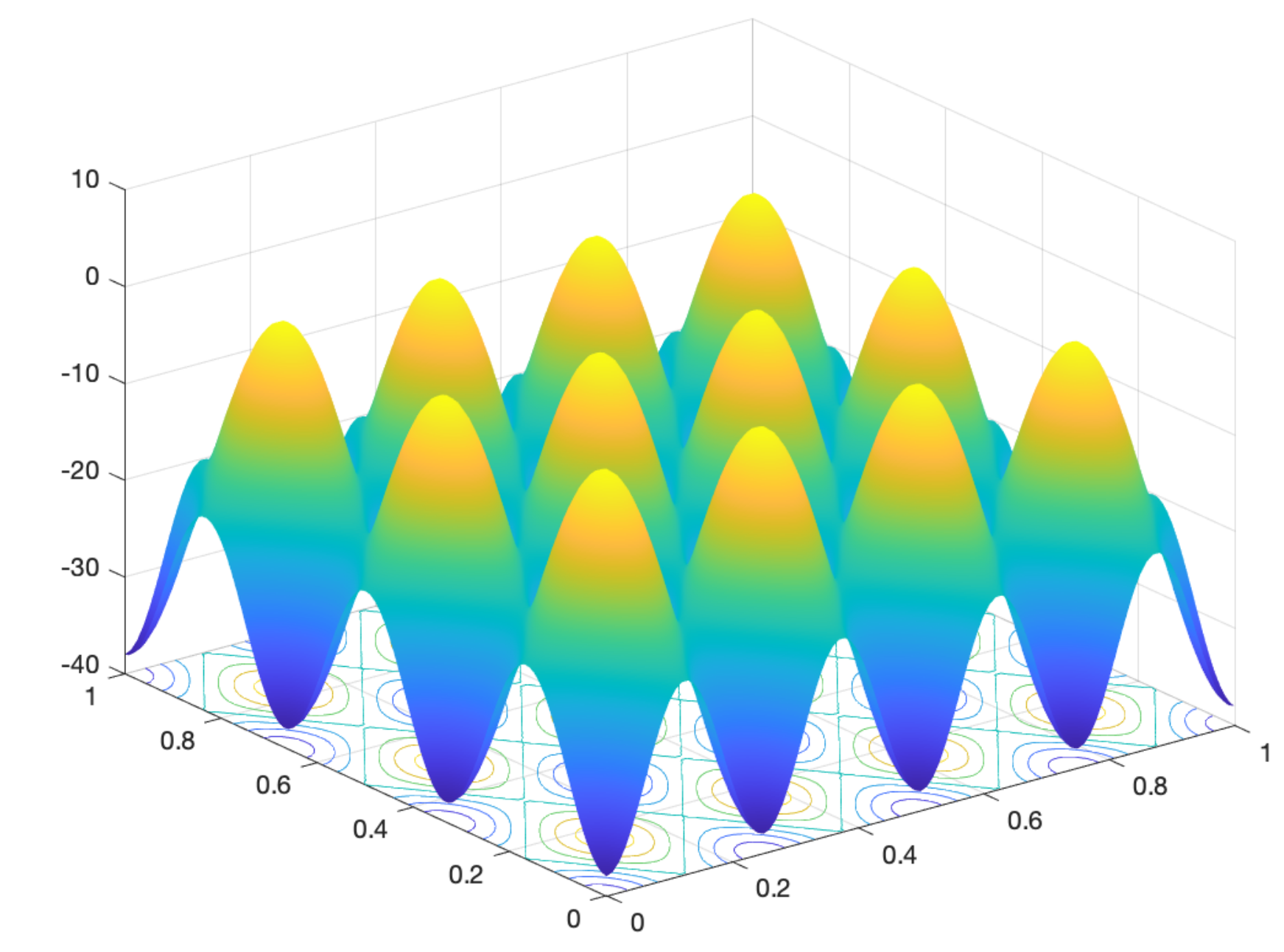}}
    \subfigure[F7 Results]{\label{fig:f7r} \includegraphics[trim=50 210 60 210,clip,width=0.23\textwidth]{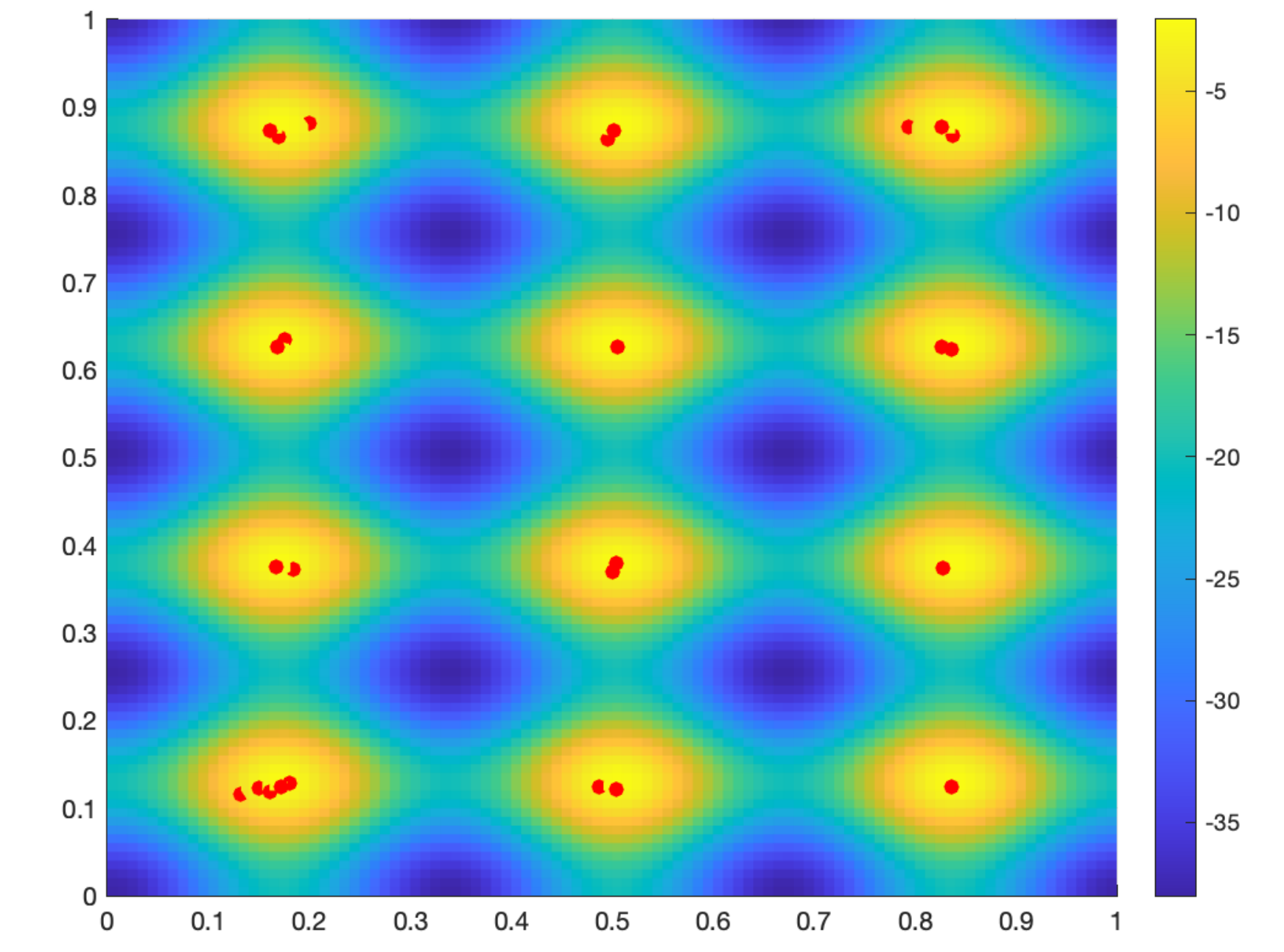}}
    \subfigure[F7 Fitness Convergence]{\label{fig:f7f} \includegraphics[trim=50 210 60 210,clip,width=0.23\textwidth]{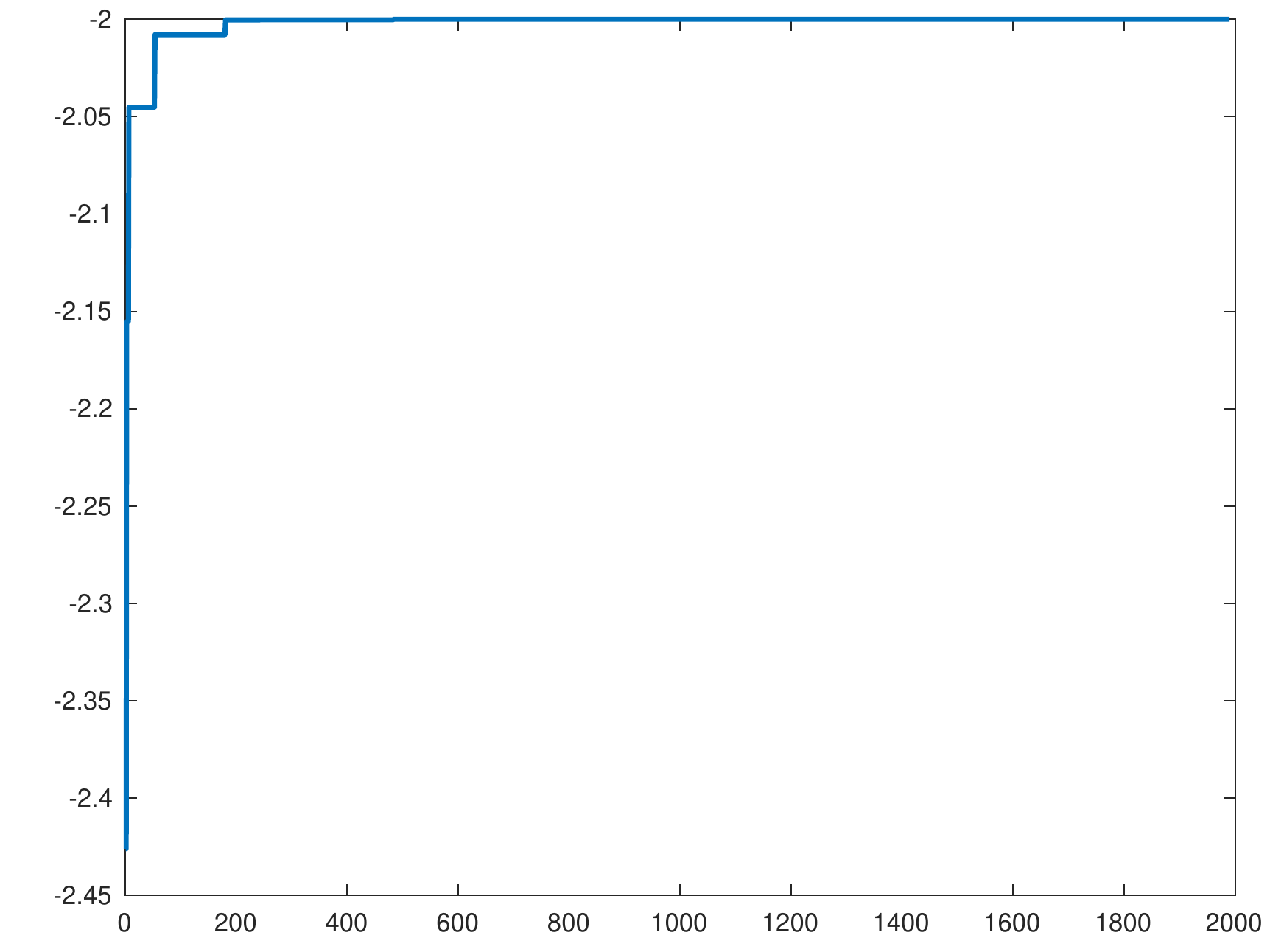}}
    \subfigure[F7 Attention Convergence]{\label{fig:f7a} \includegraphics[trim=50 210 60 210,clip,width=0.23\textwidth]{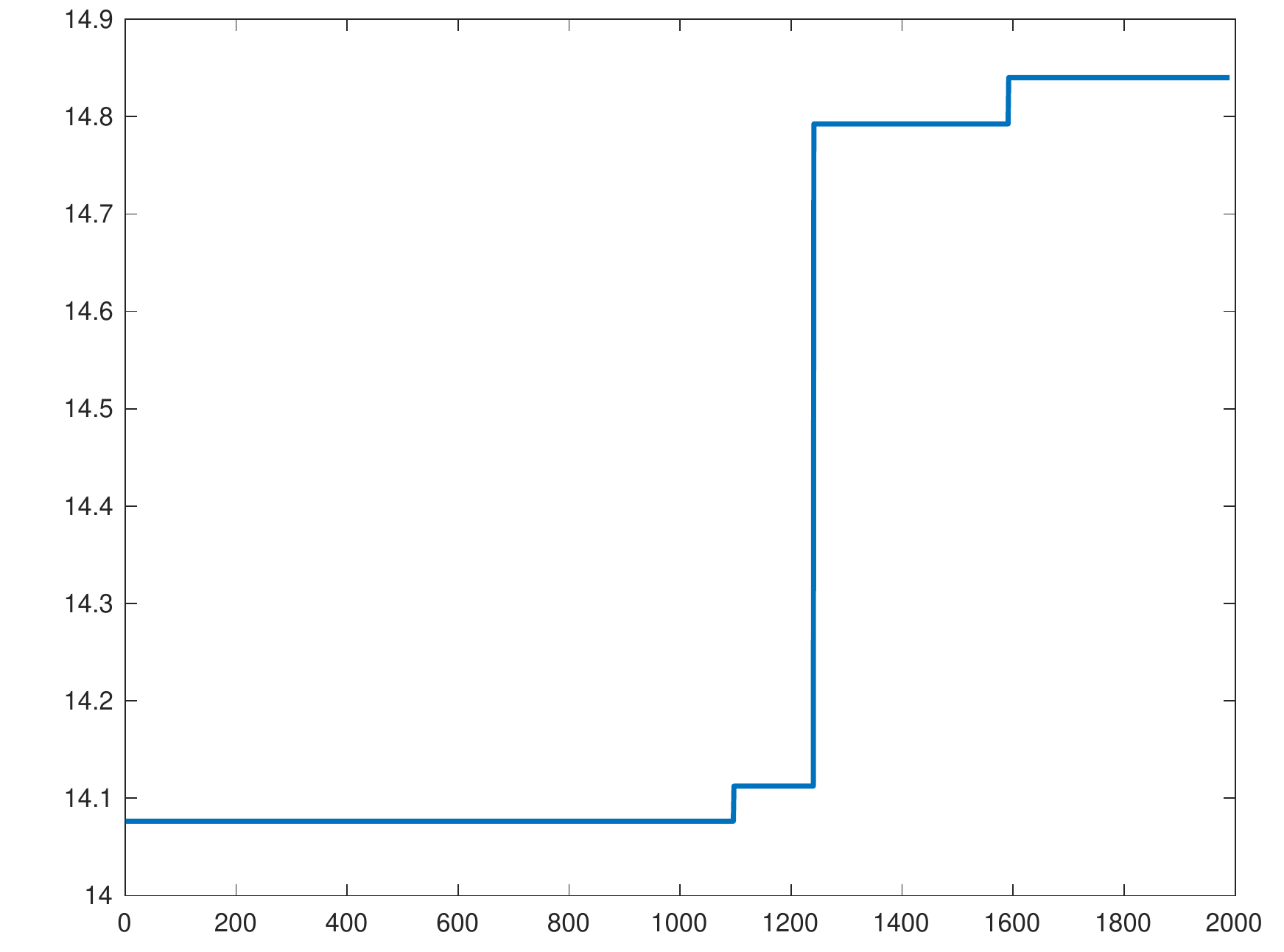}}
    \caption{Results for 2D Functions}
    \label{fig:2d}
  \end{figure*}

One set of the test results of the 2D multi-modal function is shown in Fig.\ref{fig:2d}. In the figure, each row from left to right are function plots in 3D space, solutions in top view, fitness value convergence process, and attention value convergence process, respectively.

  As same with the test results of the 1D functions, the global maximum can be found for all test functions, and the final solutions are gathered near the peak points, As shown in Fig.\ref{fig:f4r}, \ref{fig:f5r}, \ref{fig:f6r} and \ref{fig:f7r}. Since the fitness values of all individuals in the neighborhood are used in the archiving strategy (the neighborhood also includes solutions that have stored the archived list), which ensures that the global peaks can be detected. However, not all local optimal can be located. For example, in Fig.\ref{fig:f6r}, since there are too many local optimal in F6, it is impossible to detect all of them in a limited number of iterations.

  On the 2D test function, the convergence speed of the proposed method is still fast. It can be seen from Fig.\ref{fig:f4f}, \ref{fig:f5f}, \ref{fig:f6f} and \ref{fig:f7f}, the proposed method for all the four functions can quickly converge to the global optimal solution. However, since F4 and F5's landscape changes are relatively gentle, this situation causes most of the solutions have similar saliency values during iteration, which affects the convergence speed. However, although the proposed method's convergence speed on the above two functions is slightly slower than that of F6 and F7, the overall convergence speed is still relatively fast. In all 30 tests, it does not exceed 40\% of the maximum number of iterations.

  Similarly, each generation's maximum saliency values, which shown in Fig.\ref{fig:f4a}, \ref{fig:f5a}, \ref{fig:f6a} and \ref{fig:f7a}, are still changing after the fitness value has converged to the global optimum. The reasons are as described above. It indicates that the proposed algorithm is trying to find more salient solutions during rest iterations.

\section{Discussion}
The human attention mechanism is the product of long-term evolution. It can ensure that key events from the complex environment can be selected for the next processing while ignoring other insignificant parts. This mechanism has been widely developed in many fields, such as computer vision and machine learning. In fact, the attention mechanism plays an essential role in brainstorming activities.  During brainstorming, many opinions are proposed. Once some opinions attract other brainstorming participants' attention, more new ideas based on the opinions will be put forward. The BSO algorithm simulates the process of brainstorming, so introducing the attention mechanism into the algorithm will further enrich the connotation of the algorithm.

Furthermore, the multi-modal optimization problem is to extract several essential parts from a complex context. It can be analogous to multi-focal attention in the attention mechanism. Therefore, the introduction of the attention mechanism to solve multi-modal optimization problems also has excellent potential. In this paper, some exploration has been carried out on multi-modal optimization problems with attention mechanism, and preliminary results have been obtained. However, to obtain better statistical results, the relevant parameters and mechanisms need to be further refined, which is an important part of our future works.
\section{Conclusion}
This paper introduces a multi-modal optimization algorithm guided by the attention mechanism. By converting the multi-modal optimization problem's objective function into a saliency measurement function,  combined with the BSO-OS algorithm, it is performed in an ``attention objective space''. In addition, an archiving strategy and a redistribution strategy are also introduced. The former is used to preserve potential solutions, and the latter is used to increase the diversity of the population. In the archiving strategy, the solutions whose fitness value is higher than the neighborhood range are stored. It uses neighborhood individuals not only in the population of the current generation but also in the solutions in the archived list, thereby ensuring the global optimization characteristics of the algorithm. Preliminary results show that this method can locate multiple optimal in the testing benchmark functions and have excellent potential for further development. We will refine the
proposed framework in the future, and will get more statistical results with comparative studies.

\bibliographystyle{IEEEtran}
\bibliography{ABSO}

\end{document}